\documentclass[journal]{IEEEtran}

\usepackage{hyperref}
\usepackage{graphicx}
\usepackage{subfigure}
\usepackage[flushleft]{threeparttable}
\usepackage{booktabs} 
\usepackage{amsfonts}
\usepackage{multirow}
\usepackage{tabularx}
\usepackage{xcolor}
\usepackage{footnote}
\usepackage{caption}
\usepackage{textcomp, gensymb}
\usepackage{amsmath}

\usepackage{algorithm}
\usepackage{algpseudocode}

\newcolumntype{M}{>{$}c<{$}}
\newcolumntype{Z}{>{\centering\arraybackslash}X}
\newcolumntype{C}{>{\centering\arraybackslash}p}
\newcolumntype{Y}{>{\centering\arraybackslash}X}
\newcolumntype{L}{>{\raggedright\arraybackslash}p}

\newcommand{\specialcell}[2][c]{%
  \begin{tabular}[#1]{@{}c@{}}#2\end{tabular}}

\newcommand{\leftcell}[2][l]{%
  \begin{tabular}[#1]{@{}l@{}}#2\end{tabular}}

\ifCLASSINFOpdf
\else
\fi

\begin{document}
\title{Mobile Contactless Palmprint Recognition: Use of Multiscale, Multimodel Embeddings}

\author{Steven~A.~Grosz, Akash~Godbole, and~Anil~K.~Jain,~\IEEEmembership{Life~Fellow,~IEEE}
\thanks{S.A. Grosz, A. Godbole, and A.K. Jain are with the Department of Computer Science and Engineering, Michigan State University, East Lansing, MI, 48824 USA (e-mail: groszste@cse.msu.edu, godbole1@cse.msu.edu, jain@cse.msu.edu).}
}

\markboth{Journal of \LaTeX\ Class Files,~Vol.~14, No.~8, August~2015}%
{Grosz et al.: Mobile Contactless Palmprint Recognition: Use of Multiscale, Multimodel Embeddings}

\maketitle

\begin{abstract}
Contactless palmprints are comprised of both global and local discriminative features. Most prior work focuses on extracting global features or local features alone for palmprint matching, whereas this research introduces a novel framework that combines global and local features for enhanced palmprint matching accuracy. Leveraging recent advancements in deep learning, this study integrates a vision transformer (ViT) and a convolutional neural network (CNN) to extract complementary local and global features. Next, a mobile-based, end-to-end palmprint recognition system is developed, referred to as Palm-ID. On top of the ViT and CNN features, Palm-ID incorporates a palmprint enhancement module and efficient dimensionality reduction (for faster matching). Palm-ID balances the trade-off between accuracy and latency, requiring just 18ms to extract a template of size 516 bytes, which can be efficiently searched against a 10,000 palmprint gallery in 0.33ms on an AMD EPYC 7543 32-Core CPU utilizing 128-threads. Cross-database matching protocols and evaluations on large-scale operational datasets demonstrate the robustness of the proposed method, achieving a TAR of 98.06\% at FAR=0.01\% on a newly collected, time-separated dataset. To show a practical deployment of the end-to-end system, the entire recognition pipeline is embedded within a mobile device for enhanced user privacy and security.
\end{abstract}


\IEEEpeerreviewmaketitle

\section{Introduction}
\IEEEPARstart{B}{iometric} systems in operation have traditionally focused on face, fingerprint, and iris modalities, each with their own unique advantages and limitations~\cite{jainbiometrics2011}. Iris recognition gained prominence in large-scale de-duplication due to its high discriminability, but its use in commercial applications has been limited by the ease of use and relatively high cost. The face modality, on the other hand, functions as a low cost alternative, but continues to face controversy in terms of demographic bias and privacy concerns. Fingerprint usage has expanded from criminal identification to diverse commercial and government uses, but recent years have witnessed a notable shift towards contactless biometrics, spurred by the COVID-19 pandemic, where fingerprint recognition faces challenges (e.g., contact to contactless compatibility~\cite{grosz2021c2cl}, distortion~\cite{dabouei2019deep}, limited data~\cite{dong2023synthesis}, etc.). Still, face and fingerprint recognition continue to dominate the government (border crossing, civil registration), law enforcement (surveillance, latent fingerprints), commercial (access control) and personal identification (mobile payment and unlock) markets, in part due to a variety of government databases (driver license, passport, immigration, civil ID) which store vast amounts of legacy face and fingerprint images.

\begin{figure*}[t]
\begin{center}
\includegraphics[width=0.9\linewidth]{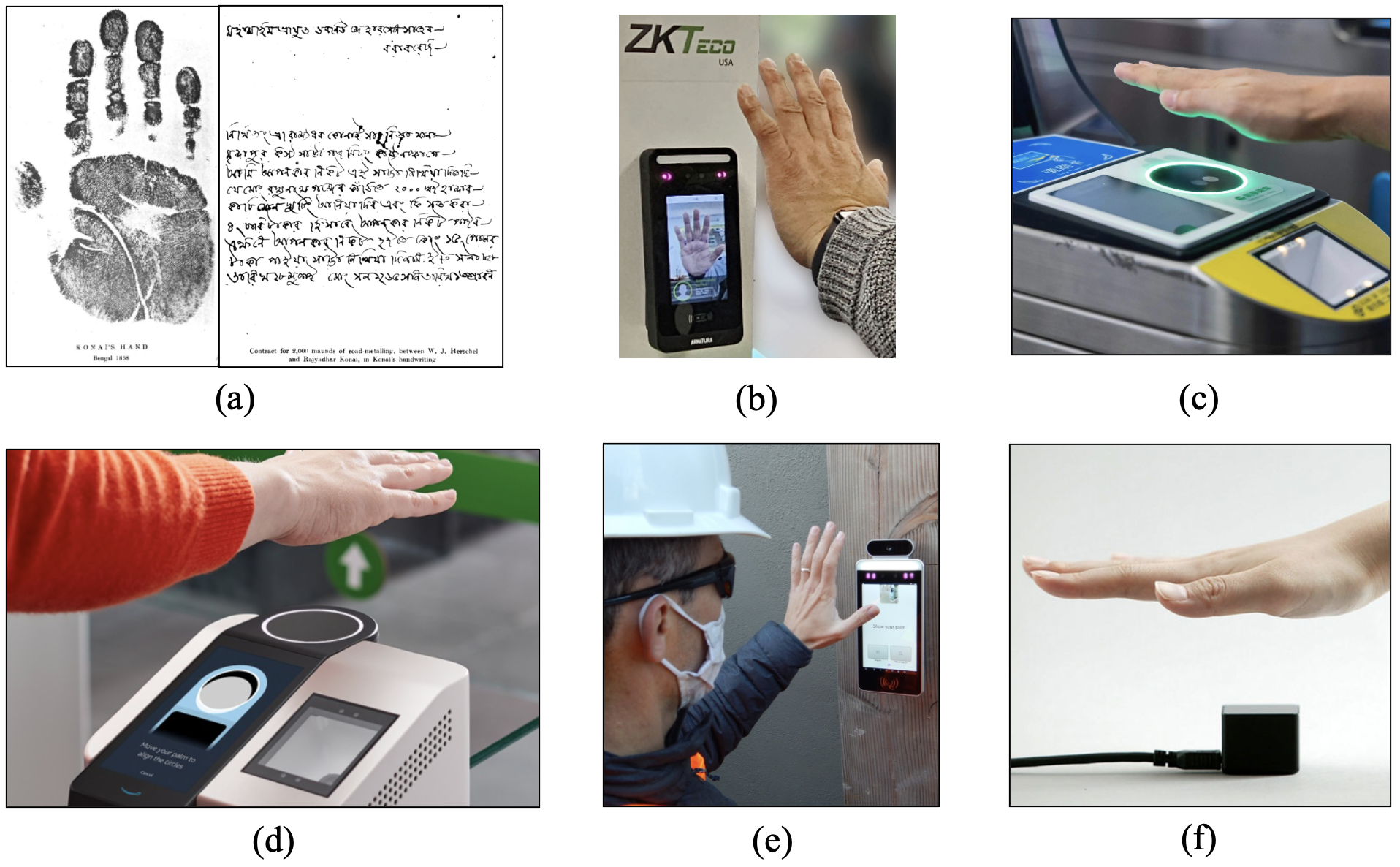}
\caption{Applications of palmprint recognition; (a) The earliest use of a palmprint impression by Sir William Herschel as a signature for a contract \cite{herschel1916origin} in the 1850s, (b) mobile contactless palmprint recognition system by ZKTeco/Armatura~\cite{armatura}, (c) contactless palmprint recognition for train and metro systems in China by Tencent \cite{Yang_2022}, (d) Amazon One system for seamless and unattended grocery shopping \cite{amazon}, (e) contactless palmprint recognition system by RedRock for employee attendance \cite{redrock}, and (f) PalmSecure palm vein recognition by Fujitsu Global \cite{fujitsu}. }
\label{fig:example_systems}
\end{center}
\vspace{-1.5em}
\end{figure*}

Nevertheless, the growing interest in contactless biometrics has prompted a surge in attention towards contactless palmprint recognition. To capture contactless palmprint images, various devices and sensing mechanisms can be used, such as direct view cameras using visible light, near-infrared cameras, and other frequency bands. Real-world biometric systems employing these mechanisms are illustrated in Figure~\ref{fig:example_systems}. Depending on the technology, captured images may depict the palmprint surface, underlying palm vein structure, or both.

For instance, the Armatura touchless palm recognition system utilizes both visible and infrared illumination to simultaneously capture surface and subsurface vein images. While fusing information from both types of images can enhance recognition accuracy, systems relying on advanced imaging for subsurface details incur higher costs. In contrast, systems using only visible spectrum palm surface images, especially those captured by widely available smartphone cameras, are more easily integrated into various applications. Consequently, there is a recent emphasis on developing accurate, cost-effective palm recognition systems using smartphone-captured images.


\subsection{Related Work}
Despite the fact that government agencies do not typically maintain palmprint databases, palmprint recognition has been an area of significant research interest in recent years due to its high amount of discriminative features (large number of minutiae, palm creases, hand geometry, etc.), ease of use, and lower privacy concerns compared to other biometric modalities. In this section, for the sake of brevity, we limit our discussion to the recent advancements in the field of contactless palmprint recognition.

Early palmprint recognition methods relied on handcrafted features extracted from a variety of local cues such as principal lines and landmark points for region of interest extraction \cite{7754902, lu2003palmprint, connie2005automated} and achieved reasonable accuracy on benchmark datasets of the time. However, many of these methods \cite{zhang2017towards, sunordinal}, operated on high resolution images (500 ppi or more), obtained from contact-based flatbed scanners collected under controlled settings (e.g., laboratory environment, low degrees of freedom due to flat imaging surface, etc.). However, contactless palmprint imaging from a smartphone is typically limited to lower resolutions (e.g., $<$ 200 ppi) and captured in a much less constrained scenario, introducing six degrees of freedom between the position of the user's palm and the capture device.

To overcome the large intra-class (intra-palm) variability introduced by contactless imaging, SOTA palm recognition systems have moved toward deep learning-based features (i.e., embeddings) for matching \cite{trabelsi2022efficient, godbole2023child}. Many of these systems extract a single, global embedding, while some (such as the BEST algorithm \cite{yulin2023best}) utilize multiple local embeddings to accumulate similarities throughout the captured palmprint image. Despite the impressive accuracy achieved by the BEST algorithm across several benchmark datasets, the speed of matching a single global embedding for each candidate image pair is a much more efficient operation (consisting of an inner product between two fixed-length vectors) compared to aggregating multiple features from several locations in the input images. However, a cascade of both global and local feature matching provides an opportunity to leverage the speed of a single, fixed length embedding with the added reliability of local matching.

\begin{table*}
    \centering
    \caption{Summary of SOTA Deep Learning Based Contactless Palmprint Recognition Literature}
    \begin{tabular}{lccc}
        \toprule
        Authors &  Study Highlight &  Limitations \\
        & (\# of unique palms) & & \\
        \toprule
        Godbole et al. \cite{godbole2023child}, 2023 & TAR=80.5\% @ FAR=0.1\% on longitudinal child palmprints  &  Large model size \\
        \midrule
        Yulin and Kumar \cite{yulin2023best}, 2023  & TAR=99.57\% @ FAR=0.1\% on CASIA Palmprint &  Train/test overlap  \\
        &  & Database using feature fusion & \\
        \midrule
        Liang et al. \cite{liang2022innovative}, 2022  & EER=0.004\% on Tongji Palmprint Database  &  Train/test set overlap \\
        & & using fusion of palmprint and palmvein & \\
        \midrule
        Zhao and Zhang \cite{zhao2020deep}, 2020 & EER=0.0052\% on CASIA and EER=0.0038\% on IITD &  Slow feature extraction and matching speed  \\
        \midrule
        Matkowski et al. \cite{matkowski2019palmprint}, 2019 & EER=0.73\% on CASIA Palmprint Database &  Train/test set overlap \\
        \bottomrule
        \end{tabular}
    \label{related_work}
\vspace{-1.0em}
\end{table*}

\subsection{Our work and contributions}
The variability in 3D hand positioning during contactless palmprint acquisition results in decreased intra-class similarity among multiple captures of the same palm. Some differences are localized, particularly near extremities, such as the lower palm region affected by thumb positioning. Therefore, relying on a single global embedding may not suffice for an accurate and robust contactless palm recognition system. Additionally, contactless palmprint images exhibit high inter-class (inter-palm) similarity due to the absence of fine-ridge details and low contrast from large standoff (distance of capture). Relying solely on local feature matching can lead to numerous false matches among visually similar palmprint images sharing common principal line structures. These challenges motivated our novel framework that combines global and local features for enhanced palmprint matching accuracy.

Recent advancements in deep learning have witnessed the rise of vision transformer (ViT) architectures~\cite{vaswani2017attention}, rivaling convolutional neural networks (CNNs) for various computer vision applications, including biometrics~\cite{tandon2022transformer, zhong2021face, delgado2023exploring}. Studies suggest that CNNs and ViTs encode different and complementary features for the same task~\cite{raghu2021vision}, motivating a fusion of both architectures for improved accuracy~\cite{grosz2022minutiae, grosz2023afr}. Building on this, we employ a fusion of a CNN and a ViT to extract a single global embedding for each palmprint image. Furthermore, multi-resolution patches are incorporated into the ViT architecture to embed varying scales of local features.

As part of our end-to-end pipeline for contactless palmprint recognition, we incorporate a novel palmprint enhancment method that utilizes domain knowledge during training to improve the downstream recognition performance. In contrast to previous palmprint enhancement approaches that rely on hand-crafted image filtering techniques, we leverage deep learning to enhance the region of interest in palmprint images. The enhancement model is trained to address degradation such as occlusion, low contrast, and noise in the form of tattoos, while emphasizing the prominence of principal lines in the enhanced palmprint image.

To demonstrate the effectiveness of our method, we adopt cross-database evaluation protocols, encompassing cross-sensor and time-separated training and test datasets. Additionally, a large-scale palmprint image database captured via mobile phone cameras in an operational setting (a charitable hosptal in Dayalbagh, India) was collected and will be made available to the research community. This database spans multiple sessions with time intervals ranging from five to thirteen months, as well as images of both adults and children (between the ages of 9-48 months).

Lastly, unlike widespread deployments by Amazon, Tencent, and Fujitsu, where templates and matching reside in a central server, Palm-ID consolidates the entire recognition process within a mobile device. This integration includes image capture, feature extraction, template generation, and matching, all while upholding user privacy and security. To accelerate processing on embedded devices, a non-linear dimensionality reduction model reduces the embedding (template) size to 516 bytes while maintaining an average matching accuracy of TAR=96.97\% at FAR=0.01\% across five test datasets, comparable to the original average accuracy of 96.96\% at a dimensionality of 772 bytes.

In summary, the key contributions of this research are:
\begin{itemize}
    \item Development of a state-of-the-art palmprint recognition pipeline using deep learning methods, which integrates a diverse set of global and local features for matching.
    \item Introduction of a novel palmprint enhancement module that leverages domain knowledge to highlight and preserve relevant features for recognition.
    \item Implementation of an efficient and accurate palmprint recognition system through a learned, non-linear dimensionality reduction model.
    \item Introduction of a simple and effective palmprint image quality estimation method for optional rejection and re-capture of poor-quality samples.
    \item Adoption of cross-database and time-separated evaluation protocols, including a newly collected database that will be made publicly available to the research community.
    \item Development of a mobile-app for every stage of the end-to-end recognition pipeline, including data collection, region of interest extraction, feature extraction, and matching; all taking place on-device. This app will be made publicly available upon publication.
\end{itemize}

\section{Methods}
Our methodology involves the development of a comprehensive palmprint recognition system designed for mobile devices, encompassing a range of features that aggregate information from multiple scales. Initially, upon image acquisition, a region of interest (ROI) is determined by estimating keypoints on the input palm image and applying a non-linear homography transformation to predefined destination points. Subsequently, a spatial transformer network, parametrized by a non-linear thin-plate-spline (TPS), precisely aligns the input ROI. Features are then extracted from the aligned ROI and matched using the cosine similarity measure between the feature vector of the input probe and the corresponding gallery feature vectors. To ensure high system throughput, a non-linear dimensionality model is employed to reduce the size of our feature vectors (i.e., embeddings) with no degradation in recognition accuracy.

\begin{figure*}[t]
\begin{center}
\includegraphics[width=\linewidth]{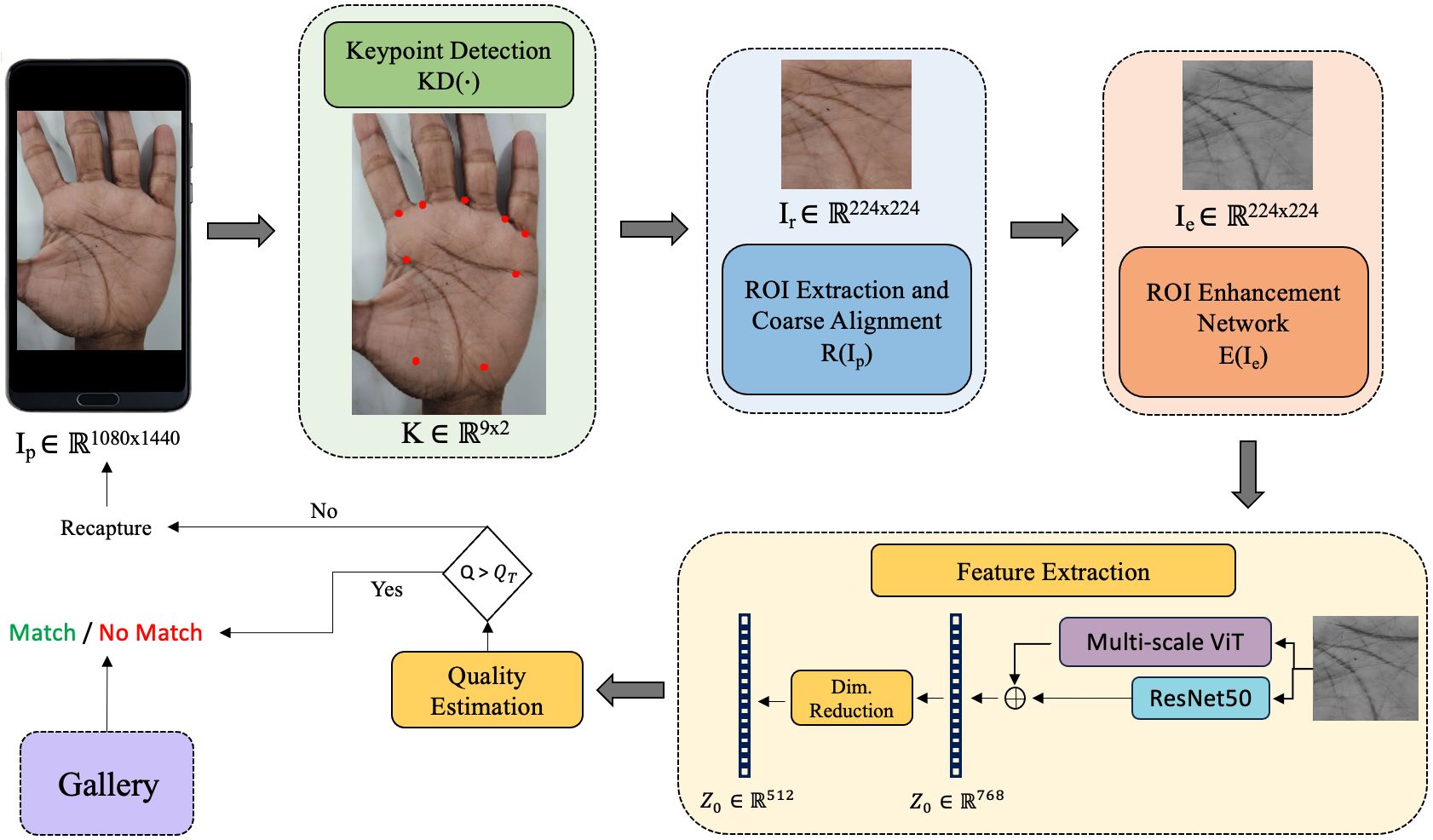}
\caption{Overview of end-to-end pipeline.}
\label{fig:overview}
\end{center}
\vspace{-1.0em}
\end{figure*}

To demonstrate the practicality of our system as an end-to-end solution for contactless palmprint recognition, each step of the pipeline is seamlessly embedded into a mobile application operating on a Samsung Galaxy S22 smartphone with a Qualcomm Snapdragon 8 Gen 1 processor with 8GB of RAM. Lastly, we introduce a Deep Neural Network (DNN)-based quality estimation method that enhances accuracy and robustness in deployment scenarios, providing the option to reject low-quality images. Subsequent subsections will delve into the detailed description of each component of our system. Figure~\ref{fig:overview} provides an overview of the entire pipeline. Each learned component shares the same training and validation datasets, which are listed in Table~\ref{databases}

\subsection{ROI Extraction}
\begin{figure*}[t]
\begin{center}
\includegraphics[width=\linewidth]{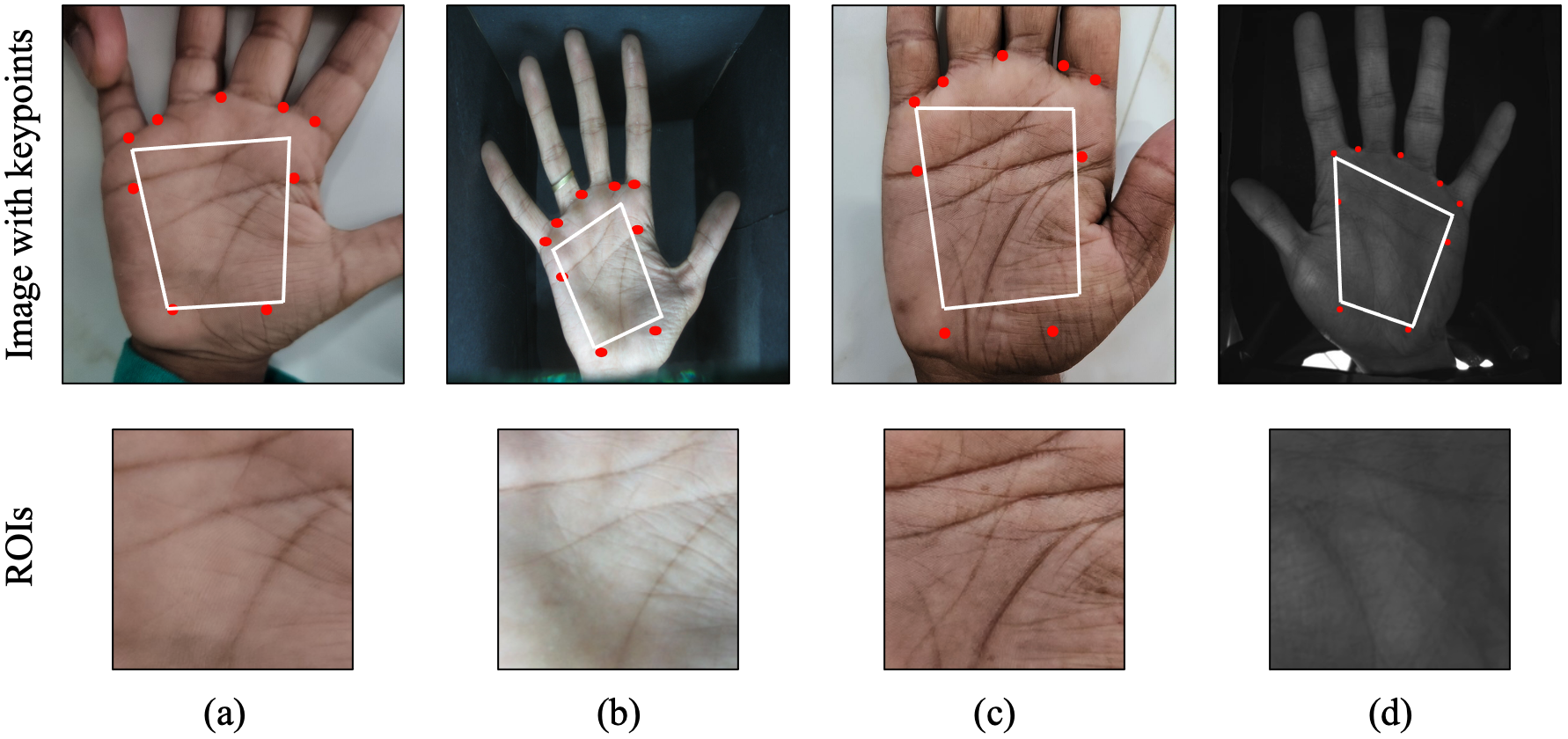}
\caption{Example keypoint predictions and corresponding ROIs from (a) MSU-CPDB-3, (b) IITD v1 \cite{kumar2008incorporating}, (c) MSU-APDB-3, and (d) CASIA Multispectral \cite{hao2008multispectral} contactless palmprint datbases.}
\label{fig:roi}
\end{center}
\vspace{-1.0em}
\end{figure*}

After successfully capturing a palmprint image, the obtained image is fed into the keypoint extraction module denoted as $K(\cdot)$. This module predicts nine keypoints positioned around the palmar boundary. Example keypoint predictions and the corresponding Region of Interest (ROI) extractions are depicted in Figure~\ref{fig:roi}. The $K(\cdot)$ module is comprised of a ResNet-50 model trained via an MSE loss using keypoints generated by a commercial-off-the-shelf (COTS) palmprint SDK from Armatura~\cite{armatura} as ground truth. Leveraging these nine keypoints, the ROI is precisely localized through a homographic transformation involving nine meticulously selected destination points in a uniform coordinate system. These destination points were hand selected to encapsulate the most discriminate and stable regions of the palm.

During training, a range of data augmentations (translation, rotation, scaling, blurring, and perspective transforms) are randomly applied to build a robust model with high accuracy, particularly on images with substantial pose variations commonly encountered in contactless palmprint images. Following the homographic transformation, an ROI with dimensions $224\times 224$ is cropped and employed as input for the enhancement network. The network is trained with the Adam~\cite{kingma2014adam} optimizer for 200 epochs with a batch size of 32, learning rate of 0.001, and weight decay of $2e^{-5}$. The final model was selected according to the highest accuracy obtained on the validation set.

\subsection{Enhancement}
In the context of contactless palmprint recognition, where the capture scenario is unconstrained, various sources of noise may be present in acquired images, potentially compromising matching performance. Notably, there is a significant likelihood of encountering very low-contrast images, arising from diverse factors such as a substantial stand-off distance between the palm and the capture device, motion blur, and over-saturation. Other degradations, including extreme perspective distortions and occlusions, may also be observed. While perspective distortions can be partially alleviated through non-linear region of interest extraction and robust feature extraction, as previously outlined, we specifically address the sensitivity to occlusions and low contrast by implementing a palmprint enhancement module through learning.

\begin{figure}[t]
\begin{center}
\includegraphics[width=\linewidth]{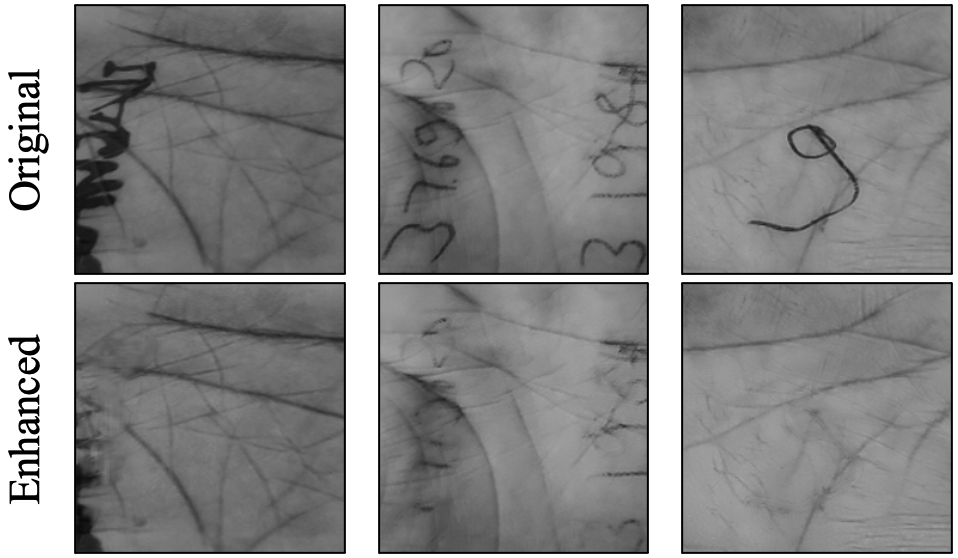}
\caption{ROIs (top row) with handwriting on the palmar surface and their enhanced counterparts by the proposed method (bottom row)}
\label{fig:enh}
\end{center}
\vspace{-1.0em}
\end{figure}

One straightforward method for improving palmprint quality involves image sharpening techniques such as high-pass filtering, Discrete Fourier Transform (Jain et al.~\cite{jain2008latent}), Gabor filters (Cappelli et al.~\cite{cappelli2012fast}), and others. The goal is to accentuate the distinctive edge features of palm creases. However, a limitation of these approaches is that the parameters of the enhancement modules are tuned to a specific dataset and in practice do not generalize to newly acquired palmprint images from different devices and environments. Thus, rather than design a hand-crafted approach, which relies heavily on correct hyper-parameterization, we turn to deep learning for palmprint image enhancement.

\begin{figure}[t]
\begin{center}
\includegraphics[width=\linewidth]{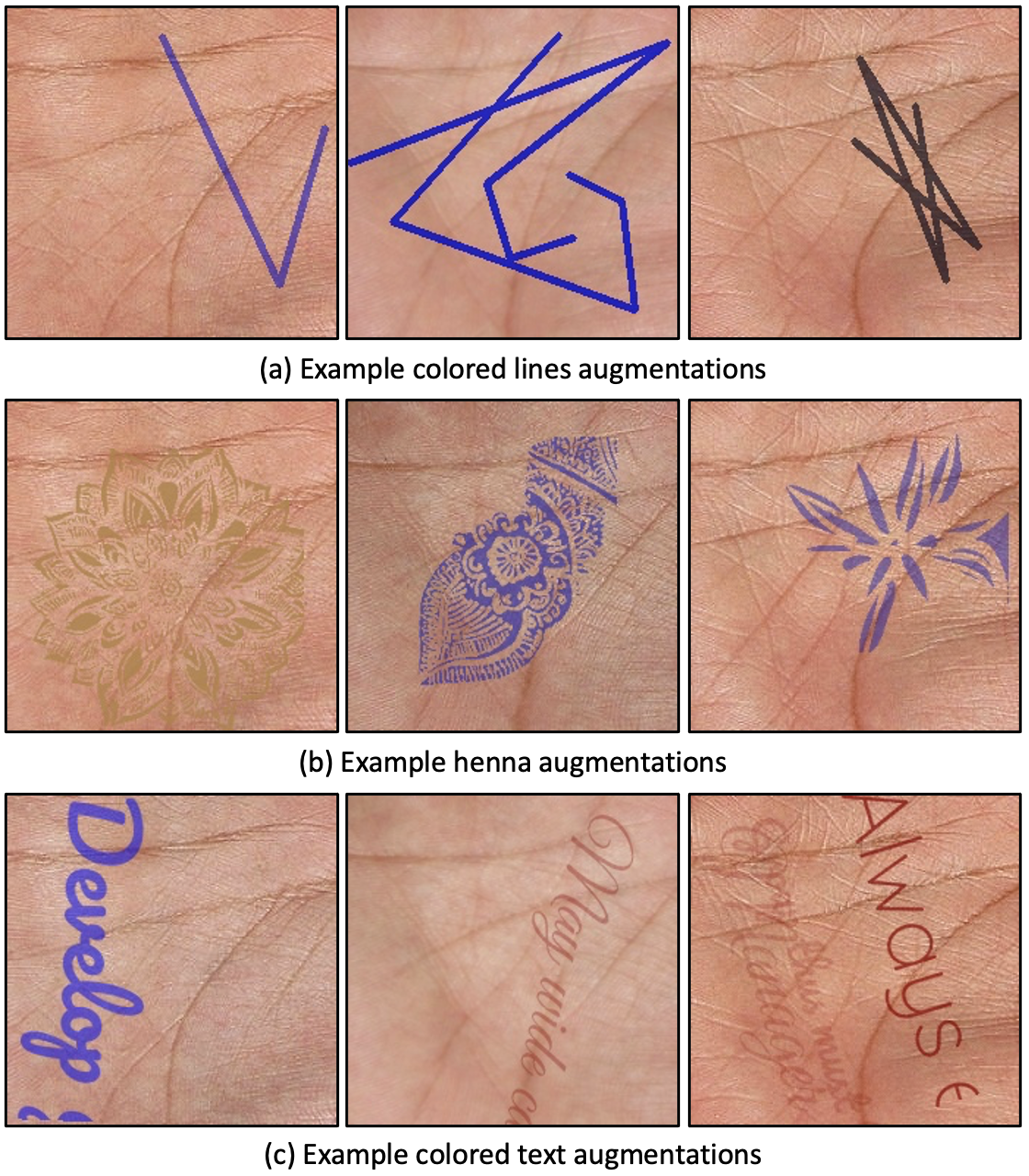}
\caption{Example augmentations for training the proposed enhancement model.}
\label{fig:enh_augs}
\end{center}
\vspace{-1.0em}
\end{figure}

Motivated by the recent application of DNNs for latent fingerprint image enhancement~\cite{grosz2023latent}, we employ the efficient SqueezeUNet architecture for palmprint enhancement. We guide the network to remove certain degradations (low contrast and occlusions) present in the palmprint images by adding a series of augmentations to the ground-truth, high-quality palmprint images and training the network to recover the original images (see Fig. \ref{fig:enh}) via an MSE loss. The augmentations that we add are specific to the contactless palmprint recognition scenario and include Gaussian blurring, random down-sampling, and overlaying text (tattoos or handwriting), colored lines, and henna-like patterns on the surface of the palm. Various examples of these augmentations are shown in Figure~\ref{fig:enh_augs}. The network is trained with the Adam optimizer for 75 epochs with a batch size of 64, learning rate of 0.001, and weight decay of $2e^{-5}$. A validation set was used during training the save the model with highest validation accuracy.

To create henna patterns (tattoos), often found on the palms of individuals in several countries especially during festivals and marriage\footnote{ https://en.wikipedia.org/wiki/Mehndi}, we employ the Stable Diffusion implementation~\cite{rombach2022high} from the Diffusers library~\cite{von-platen-etal-2022-diffusers}. Utilizing the textual prompt ``colorful henna flower sketch on white background," we generated 200 henna patterns. After a manual filtering process to eliminate poorly generated samples, we retained 143 high-quality henna patterns. These patterns are then integrated into the original palmprint ROIs to produce authentic-looking henna palmprints. The procedure involves obtaining a mask outline for each pattern, randomly resizing and cropping the mask, assigning a random RGB color, and blending the colorized mask with the original image using a random transparency level within the [0.1, 1] range, where 0 represents full transparency. 

To create random text and marker patterns, a comparable method is employed, but without utilizing Stable Diffusion. In this case, random English language sentences are generated using the Faker library in Python, employing five distinct font styles. Additionally, random line markings are generated by using the OpenCV function ``polylines," incorporating variations in size, thickness, and placement within the image.

\subsection{Feature Extraction}
We once again turn to deep learning for contactless palmprint feature extraction. Diverging from previous contactless palmprint recognition studies, we depart from a single deep network architecture approach. Instead, we leverage recent research~\cite{raghu2021vision} that advocates for the synergistic qualities of combining Vision Transformers (ViT) and Convolutional Neural Networks (CNNs). Specifically, we incorporate both the small version of ViT and the ResNet50 architectures in our method.

\begin{figure*}[t]
\begin{center}
\centering
\includegraphics[width=\linewidth]{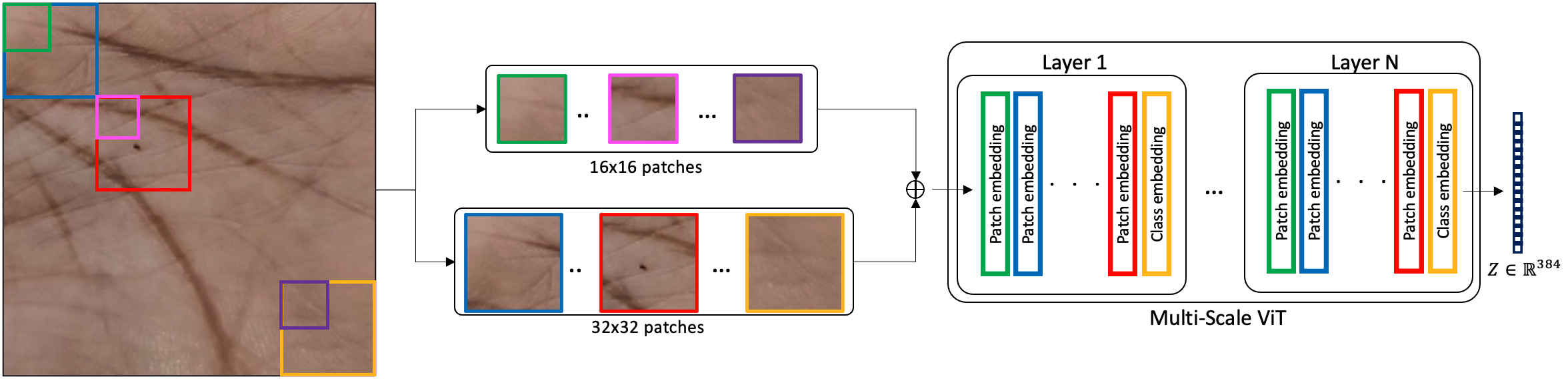}
\caption{Illustration of multi-patch resolutions used as input to ViT.}
\label{fig:multi-patch}
\end{center}
\vspace{-1.0em}
\end{figure*}

Given the variability in stand-off distance, leading to differences in scale and resolution across captured palmprint images within various databases, capturing features at diverse scales is crucial for robust contactless palmprint recognition. ViT and CNNs inherently process information at distinct scales across their networks, with ViTs aggregating global information at earlier layers compared to CNNs. To further enhance scale invariance, we encode multi-scale patches into our ViT architecture. Both our ViT and ResNet50 networks take, as input, palmprint ROI images resized to 224x224 pixels. For ViT, we use patch sizes of 16 and 32 as input, demonstrating the multi-scale patches in Figure~\ref{fig:multi-patch}. An ablation study in Section~\ref{sec:ablation} quantitatively demonstrates the advantages of incorporating both patch sizes in our ViT architecture, as well as the synergies achieved by combining CNN and ViT over using each network individually.

The ResNet50 network is trained with an Adam optimizer for 150 epochs with a batch size of 1024, learning rate of 0.001, and weight decay of $2e^{-5}$. The ViT model is first trained with a patch size of 16 using an AdamW~\cite{loshchilov2017decoupled} optimizer for 350 epochs with a batch size of 1024, learning rate of 0.001, and weight decay of $2e^{-5}$. The ViT model is then finetuned for 50 epochs using both patch sizes of 16 and 32 as input features with a learning rate of 0.0001. Both ViT and ResNet50 models are trained with the ArcFace classification loss~\cite{deng2019arcface} and the best models were saved based on highest validation accuracy.

\subsection{Matching}
Similarity scores are computed by taking the dot product between L2-normalized embeddings. To be more specific, we concatenate each normalized 384-dimensional feature embedding from Vision Transformer (ViT) and ResNet50 into a unified 768-dimensional embedding. Before the concatenation, each of the embeddings, denoted as $Z_v$ and $Z_r$, are normalized to unit length. Given that the dot product between the concatenated embeddings ranges from [0,2], we re-normalize the scores to fall within the range [0,1] via the following equation:

\begin{equation}
    s = \frac{(Z_{v}^T \cdot Z_{r}) + 2}{4}
\end{equation}

\subsection{Dimensionality Reduction}
Figure~\ref{fig:multi-patch} provides an illustration of dimensionality reduction through the DeepMDS++ model. The codebase used is adapted from~\cite{engelsma2022hers}, relying on DeepMDS~\cite{gong2019intrinsic} for the task of reducing the dimensionality of image embeddings. DeepMDS transforms the ambient representation into a minimal intrinsic space, effectively reducing dimensionality while preserving essential discriminative features.

Distinct dimensionality reduction models are trained for ViT and ResNet embeddings instead of using a single model for the concatenated embeddings. Training a single reduction model on concatenated embeddings proved challenging due to the variance in latent space between the two representations. The best results were obtained when compressing each of the individual embeddings from a dimension of 384 to 256, resulting in a concatenated embedding of dimension 512. Although experimentation with even smaller embedding sizes was conducted, we ultimately settled on 512 for our final model as it exhibited no decline in accuracy compared to the original embedding size. Experiments with other embeddings sizes are given in section~\ref{sec:efficiency}.

In order to significantly reduce the storage size of the embeddings, we employ compression by converting the precision from floats (4 bytes) to unsigned integers (1 byte). Initially, we calculate the maximum and minimum float values within the embedding and conduct a min-max normalization, ensuring the values fall within the range [0, 1]. Subsequently, each value is scaled to fit within [0, 255], and the precision is truncated from 32 bits to 8 bits by converting from float to unsigned integer. Lastly, we convert the minimum and maximum values to 16-bit floats and store them alongside the compressed embedding, resulting in a total storage requirement of $n+4$ bytes, where $n$ represents the dimension of the uncompressed embedding. During the matching process, the embeddings undergo decompression by converting them back to floats within the range [0,1]. The values are then re-normalized using the stored maximum and minimum values. Empirically, compressing and decompressing the features in this manner demonstrated no loss in accuracy throughout the experiments.


\subsection{Quality Estimation}
Assessing the quality of contactless palmprints is a crucial practical consideration for any deployed biometric system. Consequently, we have opted to integrate a quality metric into our end-to-end pipeline, providing the flexibility of an optional reject for processing palmprint images with low quality. However, deriving an objective measure of palmprint quality proves challenging due to the myriad factors influencing image quality, such as pose, contrast, and occlusions, which can impact downstream recognition tasks. Rather than devising a manually crafted metric or training a distinct quality prediction network, a task requiring accurate ground-truth labels, we leverage the L2 norm of our feature embeddings as a surrogate for image quality. This approach, as highlighted in Kim et al.~\cite{kim2022adaface} within the domain of face recognition, demonstrates a strong correlation with image quality. Through the adoption of this straightforward yet effective quality measure, we demonstrate a large decrease in the error rate of our palmprint recognition system with a small percentage of rejected samples.

Moreover, we conduct a comparative analysis, juxtaposing the L2 norm metric against two baseline quality measures - one from the Armatura SDK and the other employing the Laplacian of Gaussian (LoG) approach for image quality prediction. Our results reveal the superiority of our metric across all datasets that we have used here. Figure~\ref{fig:quality_vis} showcases examples of low and high-quality images predicted by our quality assessment method.

\subsection{Mobile Application}
To fulfill the goal of a real-world deployment of the proposed Palm-ID, we built an Android application that is capable of enrollment and search in real-time.

The app has three modes of operation which are listed below:
\begin{enumerate}
    \item Enrollment Mode: This mode enrolls the palmprint image and saves the templates on device.
    \item Authentication Mode: This mode is intended for 1:1 verification where the user inputs an ID and then proceeds to scan a palmprint to determine if it matches to the enrolled palmprint corresponding to the input ID.
    \item Search Mode: This mode searches the query palmprint template against all registered identities and returns the ID with the highest similarity score. To account for an open-set scenario, a similariy threshold is set to determine if the returned match scores are above an acceptable level to avoid false positives.
\end{enumerate}

\begin{figure}[t]
\begin{center}
\includegraphics[width=0.8\linewidth]{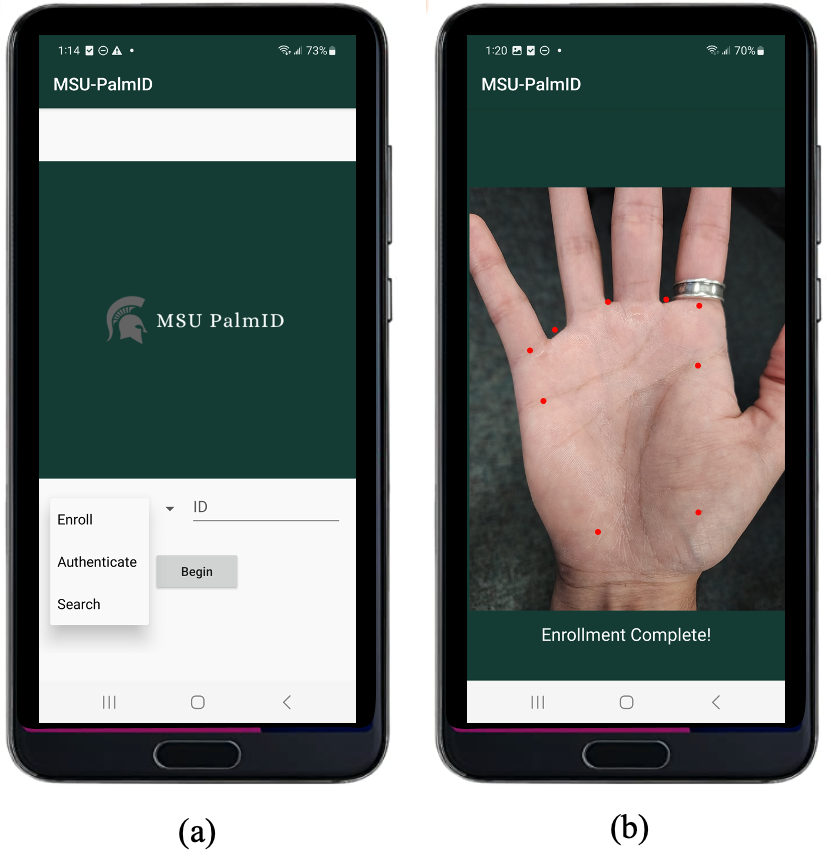}
\caption{Screenshots from our mobile application. There are 3 modes of operation in the application: i) Enrollment, ii) Authentication (1:1 comparison), and iii) Search or Identification (1:N comparison). (a) Home screen of the app with a dropdown menu to select mode of operation and (b) enrolment of a palm.}
\label{fig:app}
\end{center}
\vspace{-1.5em}
\end{figure}

Screenshots from the Palm-ID mobile application are shown in Figure~\ref{fig:app}. This app will be made publicly available to promote easier data collection efforts in the field.

\section{Experimental Results}
\label{sec:results}
In this section, we present a comprehensive evaluation of the proposed Palm-ID system against the state-of-the-art Armatura SDK and other publicly available algorithms on various benchmark datasets. We examine authentication and large scale identification scenarios, algorithm speed, template size, and also a detailed ablation study highlighting the contribution of individual components of the system.

\begin{figure*}
    \begin{center}
    \includegraphics[width=\linewidth]{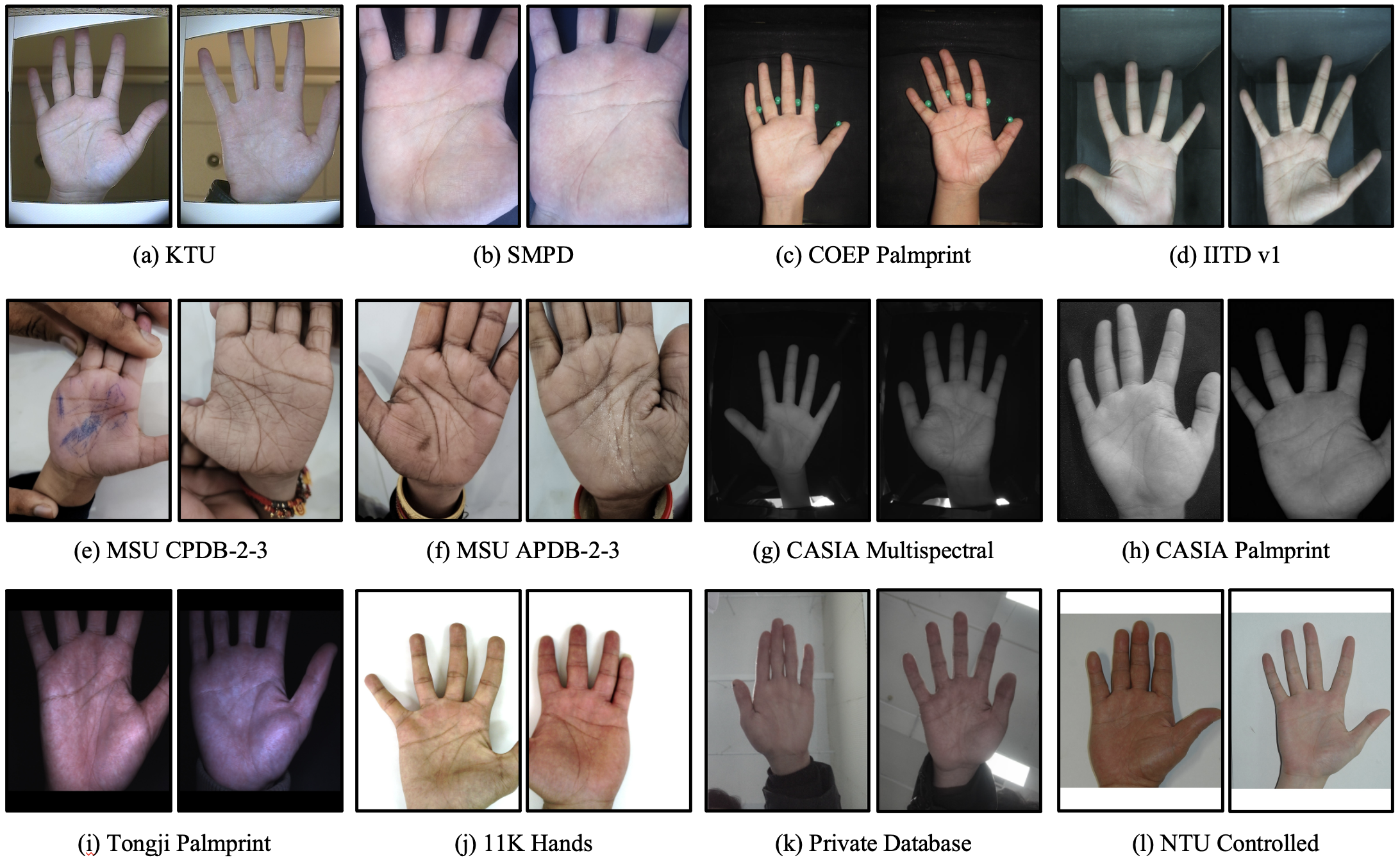}
    \caption{Two images from each database used in this paper. (a) MSU CPDB-1 \cite{godbole2023child}, (b) MSU CPDB-2 \cite{godbole2023child}, (c) MSU CPDB-3 (This Paper), (d) MSU APDB-2 \cite{godbole2023child}, (e) MSU APDB-3 (This Paper), (f) CASIA Multispectral Palmprint Database \cite{hao2008multispectral}, (g) Tongji Palmprint Database \cite{zhang2017towards}, (h) 11K Hands Database \cite{afifi201911kHands}, (i) Private Palmprint Database, (j) KTU Palmprint Database \cite{aykut2015developing}, (k) Sapienza Mobile Palmprint Database \cite{izadpanahkakhk2019novel}, (l) COEP Palmprint Database \cite{coep}, (m) IITD v1 Palmprint Database \cite{kumar2008incorporating}, (n) CASIA Palmprint Database \cite{sunordinal}, and (o) NTU Controlled Palmprint Database \cite{matkowski2019palmprint}.}
    \label{fig:collage}
    \end{center}
    \vspace{-1.0em}
\end{figure*}

\begin{table*}
\centering
\begin{threeparttable}
    \caption{Summary of Contactless Palmprint Databases used in this study.}
    \begin{tabular}{lccccc}
        \toprule
        \multicolumn{6}{c}{\textbf{Training Databases}} \\
        \midrule
        Name & \# Unique Palms & \# Images & \# Images/Palm & Capture Device & Time-separated Collection \\
        \midrule
        MSU - CPDB-1 \cite{godbole2023child} (Aug 2022) & 583 & 10,144 & 17 & Samsung Galaxy S22 & No \\
        MSU - CPDB-2 \cite{godbole2023child} (Jan 2023) & 649 & 12,508 & 19 &Samsung Galaxy S22  & No \\
        MSU - CPDB-3 (Sep 2023)\tnote{1} & 817 & 15,446 & 19 & Samsung Galaxy S22 & No \\
        MSU - CPDB-1-2 \cite{godbole2023child} (Aug 2022 - Jan 2023) & 139 & 5,090 & 37 &Samsung Galaxy S22  & Yes (5 mos.) \\
        MSU - CPDB-1-3 (Aug 2022 - Sep 2023)& 82 & 2,930 & 36 & Samsung Galaxy S22 & Yes (13 mos.) \\
        MSU - CPDB-1-2-3 (Aug 2022 - Sep 2023) & 217 & 12,535 & 58 & Samsung Galaxy S22 & Yes (13 mos.) \\
        MSU - APDB-2 \cite{godbole2023child} (Jan 2023) & 793 & 15,833 & 20 & Samsung Galaxy S22 & No \\
        MSU - APDB-3 (Sep 2023)\tnote{1} & 905 & 16,890 & 19 & Samsung Galaxy S22   & No \\
        CASIA Multispectral \cite{hao2007comparative, hao2008multispectral} (Dec 2008)& 200 & 7,200 & 36 & Custom Sensor & No \\
        Tongji \cite{zhang2017towards} (Sep 2017) & 600 & 12,000 & 20 & Custom Sensor & Yes (2 mos.) \\
        11K Hands \cite{afifi201911kHands} (Mar 2019)& 376 & 5,396 & 14 & n.a. & No \\
        \midrule
        \textbf{Total} & \textbf{5,361} & \textbf{115,972} & \textbf{-} & \textbf{-} & \textbf{-} \\
        \midrule
        \multicolumn{6}{c}{\textbf{Validation Databases}} \\
        \midrule
        SMPD \cite{izadpanahkakhk2019novel} (Jan 2019)& 91 & 1,610 & 18 & DSLR Camera & No \\
        COEP \cite{coep} (n.a.) & 146 & 1,169 & 8 & DSLR Camera & No \\
        KTU \cite{aykut2015developing} (Aug 2015)& 147 & 1,752 & 12 & Flatbed Scanner & No \\
        Proprietary Palmprint Database & 2,128 & 45,424 & 20 & n.a. & No \\
        \midrule
        \textbf{Total} & \textbf{2,512} & \textbf{49,955} & \textbf{-} & \textbf{-} & \textbf{-} \\
        \midrule
        \multicolumn{6}{c}{\textbf{Test Databases}} \\
        \midrule
        Name & \# unique palms & \# images & \# images/palm & Capture Device & Time-Separated collection \\
        \midrule
        MSU - CPDB-2-3 (Jan 2023 - Sep 2023)\tnote{1} & 258 & 10,031 & 39 & Samsung Galaxy S22 & Yes (7 mos.) \\
        MSU - APDB-2-3  (Jan 2023 - Sep 2023)\tnote{1} & 406 & 15,424 & 38 & Samsung Galaxy S22 & Yes (7 mos.) \\
        CASIA Palmprint Database \cite{sunordinal} (July 2005)& 620 & 5,502 & 9 & Custom Sensor& No \\
        IITD v1 \cite{kumar2008incorporating, kumarpersonal} (Dec 2008)& 460 & 2,601 & 6 & n.a. & No \\
        NTU Controlled \cite{matkowski2019palmprint} (Nov 2019)& 655 & 2,478 & 4 & DSLR Camera & No \\
        \midrule
        \textbf{Total} & \textbf{2,399} & \textbf{36,036} & \textbf{-} & \textbf{-} & \textbf{-} \\
        \bottomrule
        \end{tabular}
    \begin{tablenotes}
    \item[1] Will be released after the paper is accepted for publication.
    \end{tablenotes}
    \label{databases}
\end{threeparttable}
\vspace{-0.5em}
\end{table*}


\subsection{Databases}
\subsubsection{Publicly Available Palmprint Databases}
First, we give a description of all the publicly available palmprint recognition datasets used in the paper, as well as an account of our newly collected, time-separated palmprint databases. The publicly available databases used in this paper are listed in Table~\ref{databases}, along with the number of unique palms, total number of images, average number of images per palm, capture device, and time-gap between subsequent images. The table is divided into separate train, validation, and test categories indicating which databases were used to learn the parameters of each component of Palm-ID and which were used to evaluate the performance compared to the baseline methods. Note, unlike many previous palmprint recognition papers, all test databases are strictly reserved for evaluation and no images/identities have overlap with any of the training or validation datasets.

\subsubsection{MSU PalmPrint Database}
As part of our ongoing efforts in contactless palmprint recognition, we conducted three data collection sessions at Saran Ashram Hospital, Dayalbagh, India. The primary motivations for this collection initiative were twofold: i.) to amass a substantial database of infant palmprints for investigating the feasibility of infant palmprint recognition, and ii.) to overcome the limitation of many existing contactless palmprint databases by capturing images of the same palm identities after significant time intervals (e.g., 12 months apart). The initial session, held in August 2022, is comprised exclusively with child palmprints. The second session, conducted in January 2023, includes both child and adult palms, while the third session in September 2023 again features both child and adult palms. Upon publication, this database will be made publicly available to the research community. 

Images in this database are categorized based on the age of the participant (child vs. adult) and the session number (1, 2, or 3) during which the images were collected. Moreover, images of repeat subjects are segregated into distinct sets, ensuring their separation from individual collection images. Consequently, the complete database is divided into 10 disjoint subsets, enumerated as follows:

\begin{enumerate}
    \item CPDB-1: Child palms collected in session 1 only.
    \item CPDB-2: Child palms collected in session 2 only.
    \item CPDB-3: Child palms collected in session 3 only.
    \item CPDB-1-2: Same child palms collected in session 1 and session 2, but not in session 3.
    \item CPDB-1-3: Same child palms collected in session 1 and session 3, but not in session 2.
    \item CPDB-2-3: Same child palms collected in session 2 and session 3, but not in session 1.
    \item CPDB-1-2-3: Same child palms collected in session 1, session 2, and session 3.
    \item APDB-2: Adult palms collected in session 2 only.
    \item APDB-3: Adult palms collected in session 3 only.
    \item APDB-2-3: Same adult palms collected in session 2 and session 3.
\end{enumerate}





\subsection{Authentication Results}
We present authentication results on three publicly available contactless palmprint databases: i) CASIA Palmprint Image Database, ii) IITD v1 Palmprint Database, and iii) NTU Controlled Contactless Palmprint Database v1 (NTU-CP-v1). Additionally, we assess performance on two time-separated extensions, APDB-2-3 and CPDB-2-3, of our newly collected MSU Palmprint databases. The authentication performance of our proposed model is compared to three state-of-the-art methods: Godbole et al. \cite{godbole2023child}, Matkowski et al. \cite{matkowski2019palmprint}, and a COTS palmprint SDK by Armatura. We directly use the publicly available SDK for the Matkowski et al. method, rather than re-implement the algorithm ourselves. Similarly, we use the Armatura SDK out-of-the-box, without knowledge of the matching algorithm or training databases used to develop it. For Godbole et al., we re-train the network on the same training and validation set as the proposed Palm-ID method. Other baseline methods have either not made their code publicly available or use a percentage of the test databases for training, whereas we have reserved the entirety of each test database for evaluation.

Our model achieves a True Acceptance Rate (TAR) of 99.08\% at a False Acceptance Rate (FAR) of 0.01\% on NTU-CP-v1, surpassing the next best performance of 98.94\% by Armatura. Furthermore, our method demonstrates competitive TARs of 99.53\% on CASIA and 99.93\% on IITD v1, both at FAR = 0.01\%, compared to 100\% and 99.72\% by Armatura. For the time-separated datasets APDB-2-3 and CPDB-2-3, the proposed model significantly outperforms all other methods in our comparison, showcasing its efficacy in less-constrained and longitudinal recognition scenarios.

To enhance recognition system accuracy further, we perform score-level fusion of our model and Armatura. We assign a weight of 0.7 to the similarity score generated by our method and a weight of 0.3 to the similarity score from Armatura. The improvement in performance underscores the complementary nature of features extracted by both models and suggest potential for future improvements in our approach. The detailed authentication results are provided in Table~\ref{tab:authentication}.

\begin{table}
\begin{threeparttable}
\centering
\caption{Authentication results, reported as TAR (\%) at FAR=0.01\%.}
\begin{tabular}{lccccc}
\toprule
\textbf{Model} & \textbf{CASIA} & \textbf{IITD} & \specialcell{\textbf{NTU-CP-}\\\textbf{v1}} & \specialcell{\textbf{APDB-}\\\textbf{2-3}} & \specialcell{\textbf{CPDB-}\\\textbf{2-3}}\\
\toprule
\leftcell{Godbole\\et al.~\cite{godbole2023child}} & 99.5 & 99.8 & 98.76 & 97.14 & 85.35 \\
\midrule
\leftcell{Matkowski\\et al. \cite{matkowski2019palmprint}} & 98.27 & 98.9 & 98.91 & 95.61 & 81.3 \\
\midrule
Armatura \cite{armatura} & 99.72 & 100 & 98.94 & 97.78 & 84.03 \\
\midrule
Proposed & 99.53 & 99.95 & 99.08 & 	98.06 & 88.24 \\
\midrule
\leftcell{Proposed +\\Armatura\tnote{1}} & 99.88 & 100 & 99.36 & 98.56 & 91.51 \\
\bottomrule
\end{tabular}
\label{tab:authentication}
\begin{tablenotes}
\item[1] Score level fusion with weights of 0.7 and 0.3 for the proposed system and Armatura, respectively.
\end{tablenotes}
\end{threeparttable}
\end{table}

\subsection{Identification Results}
Open-set search scenarios may incur two types of recognition errors. Firstly, in a nonmated search, the matching algorithm may return an identity from the gallery with a similarity score above an acceptance threshold which does not correspond to the correct identity of the probe subject, referred to as Type I error or false positive. Secondly, in a mated search, the algorithm may return an incorrect identity from the gallery instead of the correct enrolled identity, resulting in a Type II error or false negative. The False Positive Identification Rate (FPIR), representing the Type I error rate, quantifies the fraction of nonmated searches where enrolled identities are inaccurately returned at or above the specified threshold (T). On the other hand, the False Negative Identification Rate (FNIR), corresponding to the Type II error rate, measures the fraction of mated searches where the enrolled mate falls outside the top R ranks or the comparison score of the correct mate is below T. FPIR and FNIR rely on both the size of the enrolled gallery (N) and the number of top candidates considered (R). For our experiments we use N=4,377 and R=1. Table~\ref{tab:open-set} compares the FNIR (\%) at FPIR=1\% of the proposed Palm-ID matcher against the baseline method proposed in \cite{godbole2023child}. Across all five evaluation datasets, Palm-ID signifcantly outperforms the baseline, returning an average FNIR of 4.42\% compared to 32.23\%.


For closed-set identification, results are usually reported in terms of rank R retrieval rate. This accuracy metric signifies the fraction of mated searches where the enrolled mate is at rank R or better, irrespective of the comparison score. Rank-1 retrieval rate of the proposed Palm-ID matcher is compared against the baseline method of Godbole et al. in Table~\ref{tab:closed-set}. Palm-ID obtained an average rank-1 retrieval rate of 98.24\%, compared to Godbole et al. which obtained 95.94\%.

\begin{table}
\centering
\caption{Closed-set identification results with a gallery of 4,377 unique palms, reported as rank-1 retrieval rate (\%).}
\begin{tabular}{lccccc}
\toprule
\textbf{Model} & \textbf{CASIA} & \textbf{IITD} & \specialcell{\textbf{NTU-CP-}\\\textbf{v1}} & \specialcell{\textbf{APDB-}\\\textbf{2-3}} & \specialcell{\textbf{CPDB-}\\\textbf{2-3}}\\
\toprule
\leftcell{Godbole\\et al.~\cite{godbole2023child}} & 99.30 & 99.72 & 98.57 & 97.23 & 84.90\\
\midrule
Proposed & 99.71 & 100.00 & 99.29 & 98.10 & 94.11 \\
\bottomrule
\end{tabular}
\label{tab:closed-set}
\end{table}
\vspace{-0.5em}

\begin{table}
\centering
\caption{Open-set identification results with a gallery of 4,377 unique palms, using FNIR (\%) at FPIR=1\%.}
\begin{tabular}{lccccc}
\toprule
\textbf{Model} & \textbf{CASIA} & \textbf{IITD} & \specialcell{\textbf{NTU-CP-}\\\textbf{v1}} & \specialcell{\textbf{APDB-}\\\textbf{2-3}} & \specialcell{\textbf{CPDB-}\\\textbf{2-3}}\\
\toprule
\leftcell{Godbole\\et al.~\cite{godbole2023child}} & 24.79 & 1.40 & 94.77 & 6.04 & 34.13 \\
\midrule
Proposed & 0.78 & 0.19 & 3.02 & 3.32 & 14.80 \\
\bottomrule
\end{tabular}
\label{tab:open-set}
\vspace{-1.0em}
\end{table}

\subsection{Computational Efficiency}
\label{sec:efficiency}
In addition to accuracy, practical considerations, such as model latency and template size, are important factors for real-world deployment. Table~\ref{tab:efficiency} provides a comparison of these efficiency metrics for our model against other methods. Our model excels in template extraction and large-scale search efficiency, due to the proposed dimensionality reduction and feature compression techniques. The detailed tradeoff between efficiency and accuracy is given in Table~\ref{tab:embed_dim}, illustrating the performance variations as the embedding size of the proposed model changes. The results indicate that reducing the embedding size by a factor of four (from 772 bytes to 196 bytes) only decreases the average TAR by 0.51 percentage points.

\begin{table*}
\centering
\caption{TAR at 0.01\% FAR across several datasets as template (embedding) size varies.}
\begin{tabular}{ccccccc}
\toprule
\textbf{\specialcell{Template Size}}        & \textbf{CASIA} & \textbf{IITD}  & \textbf{NTU-CP-v1} & \textbf{APDB-2-3} & \textbf{CPDB-2-3} & \textbf{Average} \\
\toprule
772 bytes & \textbf{99.53\%}   & \textbf{99.97\%} & 99.05\%     & 98.05\%    & 88.22\%    & 96.96\% \\
\midrule
644 bytes & \textbf{99.53\%}   & \textbf{99.97\%} & 99.02\%     & \textbf{98.06\%}    & 88.18\%    & 96.95\% \\
\midrule
516 bytes & \textbf{99.53\%}   & 99.95\% & \textbf{99.08\%}     & \textbf{98.06\%}    & \textbf{88.24\%}    &\textbf{ 96.97\%} \\
\midrule
452 bytes & 99.52\%   & 99.93\% & \textbf{99.08\%}     & 98.05\%    & 88.10\%    & 96.94\% \\
\midrule
388 bytes & 99.51\%   & 99.92\% & 98.96\%     & 98.05\%    & 88.10\%    & 96.91\% \\
\midrule
292 bytes & 99.51\%   & 99.95\% & 98.96\%     & 98.00\%    & 88.08\%    & 96.90\% \\
\midrule
196 bytes & 99.48\%   & 99.92\% & 98.88\%     & 97.89\%    & 86.10\%    & 96.45\% \\
\bottomrule
\end{tabular}
\label{tab:embed_dim}
\end{table*}

\begin{table}
\centering
\caption{Efficiency Comparison.}
\begin{tabular}{lcccc}
\toprule
\textbf{Model} & \textbf{\specialcell{\# Params.\\(M)}} & \textbf{\specialcell{Template\\Size\\(bytes)}} & \textbf{\specialcell{Template\\Extraction\\time (ms)}} & \textbf{\specialcell{1:10,000\\comparison\\latency (ms)}} \\
\toprule
\leftcell{Godbole\\et al.~\cite{godbole2023child}} & 	89.19 & 	3080 & 	9.08 & 	0.84 \\
\midrule
Armatura~\cite{armatura} & 	N/A & 	544 & 	$<$60 & $<$10\\
\midrule
Proposed & 	76.04 & 	516 & 	18.0 & 	0.33 \\
\bottomrule
\end{tabular}
\vspace{-0.5em}
\label{tab:efficiency}
\end{table}

\subsection{Ablation Analysis}
\label{sec:ablation}
For a detailed examination of the contribution of each individual component in our Palm-ID system, we present a comprehensive ablation study in Table~\ref{tab:ablation}. A comparison between rows 1 and 2 highlights a significant improvement with the incorporation of multi-scale patches in the ViT architecture compared to a single patch size of 16, thus advocating for the utilization of multi-resolution local features for robust palmprint recognition. Additionally, the fusion with ResNet50 results in further improvements in three out of the five datasets. Lastly, it is observed that the inclusion of the learned enhancement model played a pivotal role in boosting performance, particularly on lower-quality, time-separated datasets (APDB-2-3 and CPDB-2-3), along with marginal gains for CASIA and NTU-CP-v1 datasets.

\begin{table*}
\centering
\caption{Ablation study as various modules of the proposed system are incorporated, reported as TAR (\%) at FAR=0.01\%.}
\begin{tabular}{cccc|ccccc}
\noalign{\hrule height 1.0pt}
\textbf{ViT} & \textbf{Multi-scale Patches} & \textbf{ResNet50} & \textbf{Enhancement} & \textbf{CASIA} & \textbf{IITD} & \textbf{NTU-CP-v1} & \textbf{APDB-2-3} & \textbf{CPDB-2-3}\\
\noalign{\hrule height 1.0pt}
\checkmark & & & & 99.25 & 99.57 & 97.28 & 96.70 & 82.10 \\
\noalign{\hrule height 0.5pt}
\checkmark & \checkmark & & & 99.51 & 99.75 & 97.90 & 97.53 & 84.19 \\
\noalign{\hrule height 0.5pt}
 & & \checkmark & & 99.18 & 99.90 & 98.23 & 96.31 & 83.02 \\
\noalign{\hrule height 0.5pt}
\checkmark & \checkmark & \checkmark & & 99.51 & \textbf{99.97} & 98.88 & 97.40 & 87.67 \\
\noalign{\hrule height 0.5pt}
\checkmark & \checkmark & \checkmark & \checkmark & \textbf{99.53} & \textbf{99.97} & \textbf{99.05} & \textbf{98.05} & \textbf{88.22} \\
\noalign{\hrule height 1.0pt}
\end{tabular}
\label{tab:ablation}
\end{table*}

\section{Discussion}

\subsection{Failure Case Analysis}
\begin{figure}[t]
\begin{center}
\includegraphics[width=\linewidth]{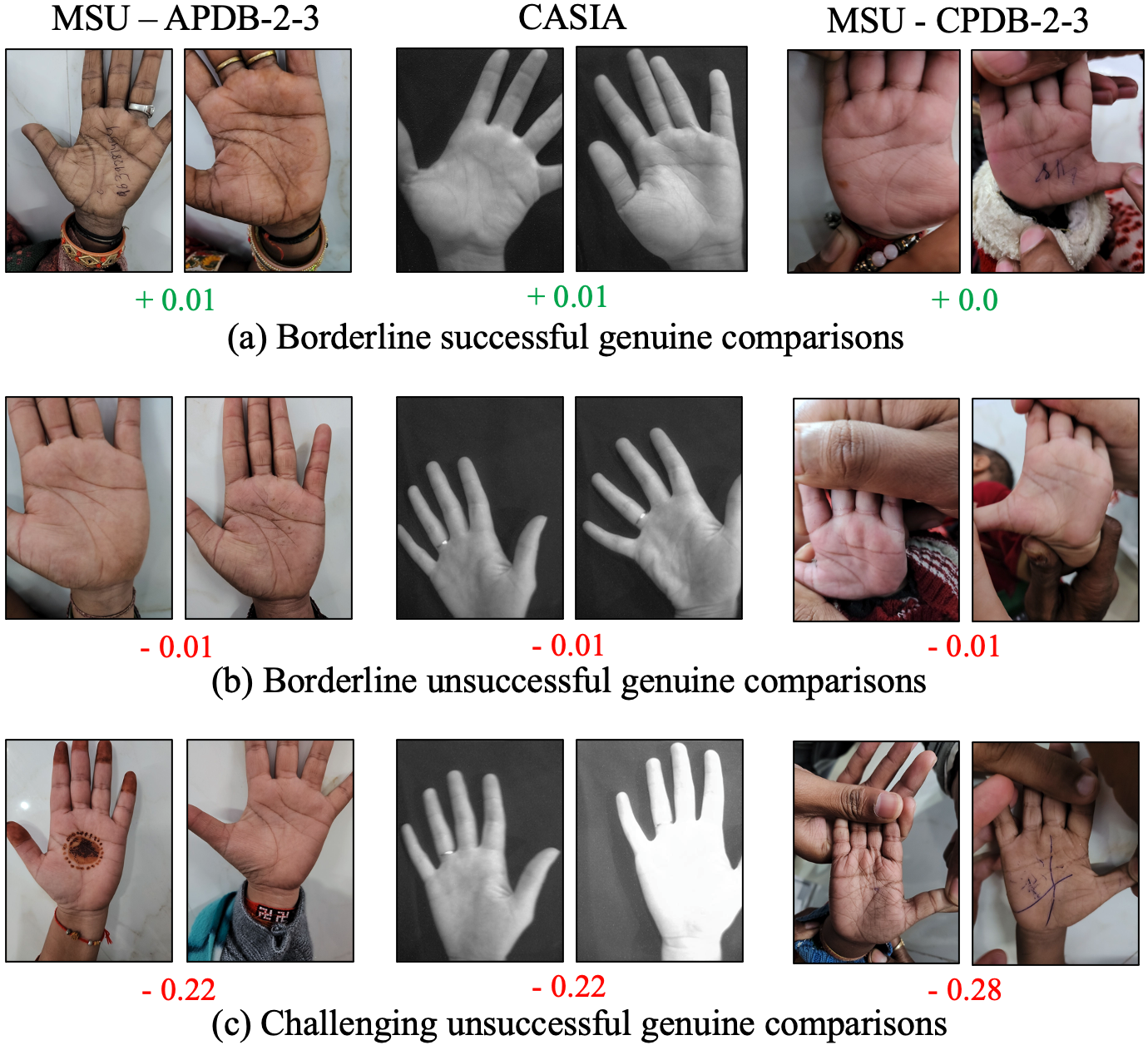}
\caption{Example success and failure cases of our model. (a) successful genuine matches which were near the match score threshold, (b) unsuccessful genuine matches where the score was near the threshold, and (c) unsuccessful genuine matches that were far from the match threshold. The number below each pair indicates the distance from the match threshold. For each row, the first pair of images are from the MSU APDB-2-3 database, the second pair is from the CASIA database, and the third pair is from the MSU CPDB-2-3 database.}
\label{fig:successes_and_failures}
\vspace{-1.5em}
\end{center}
\end{figure}

Figure~\ref{fig:successes_and_failures} illustrates both successful and unsuccessful comparisons by our method. Depicted in row (a) are some correctly matched genuine pairs that are in close proximity to the match threshold, suggesting their challenging nature for our model. Nevertheless, our system appears capable of effectively handling numerous challenging scenarios, including those with slight occlusion, blurriness, and perspective distortions. In the middle row (b) of Figure~\ref{fig:successes_and_failures}, are some failure cases representing genuine pairs that failed to match by falling just below the match threshold. Finally, row (c) showcases examples of genuine comparisons that were incorrectly classified well below the match threshold. These samples contain over-saturated images and occlusions which are challenging to be matched.

\subsection{Palmprint Quality Prediction}
\begin{figure*}[t]
\begin{center}
\includegraphics[width=\linewidth]{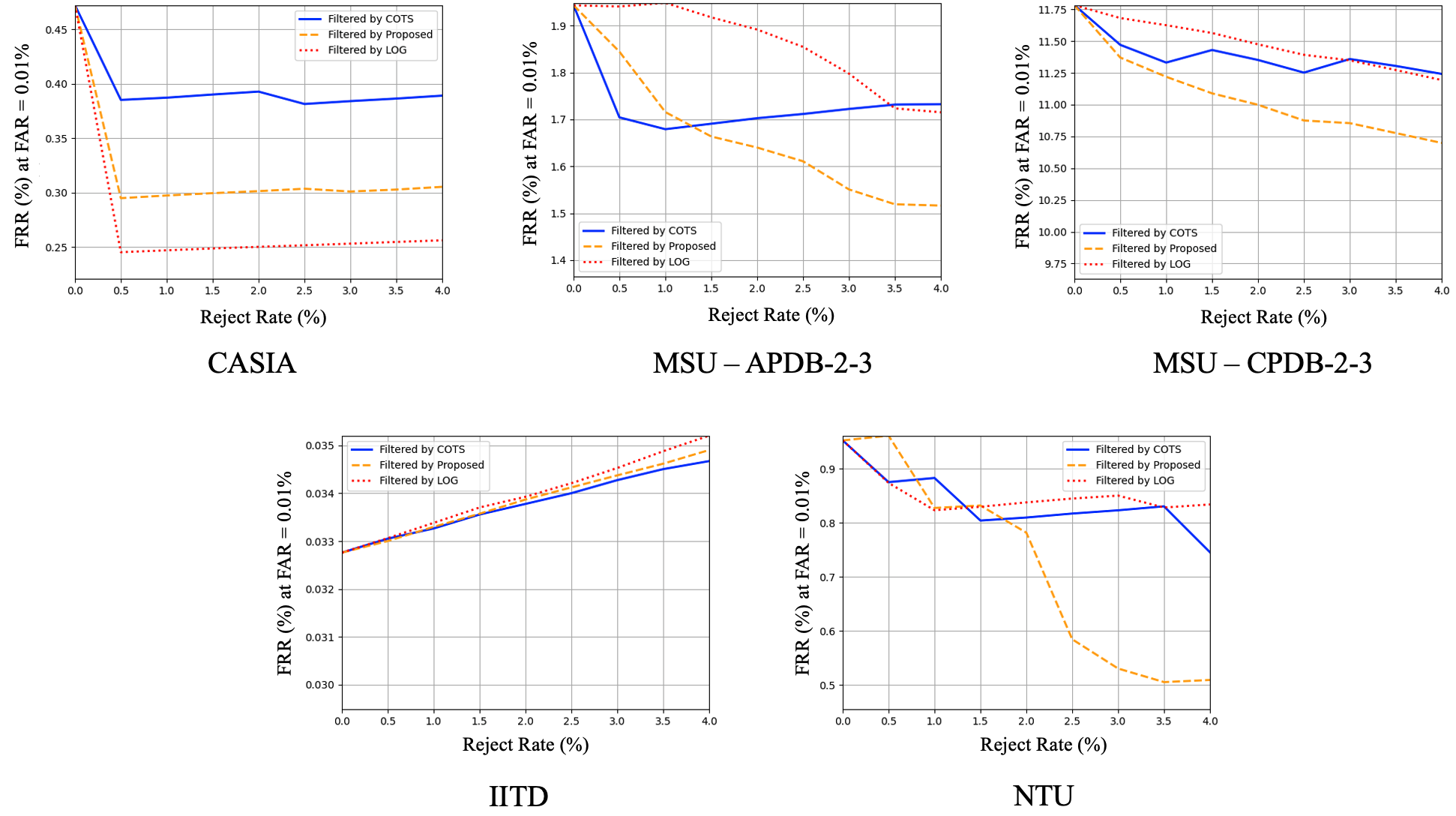}
\caption{Error-reject curves when using Armatura's SDK quality metric (blue, solid line), variance of LoG metric (red, dotted line), and the proposed quality metric for filtering (orange, dashed line).}
\label{fig:ert}
\end{center}
\vspace{-1.5em}
\end{figure*}

\begin{figure*}[t]
\begin{center}
\includegraphics[width=\linewidth]{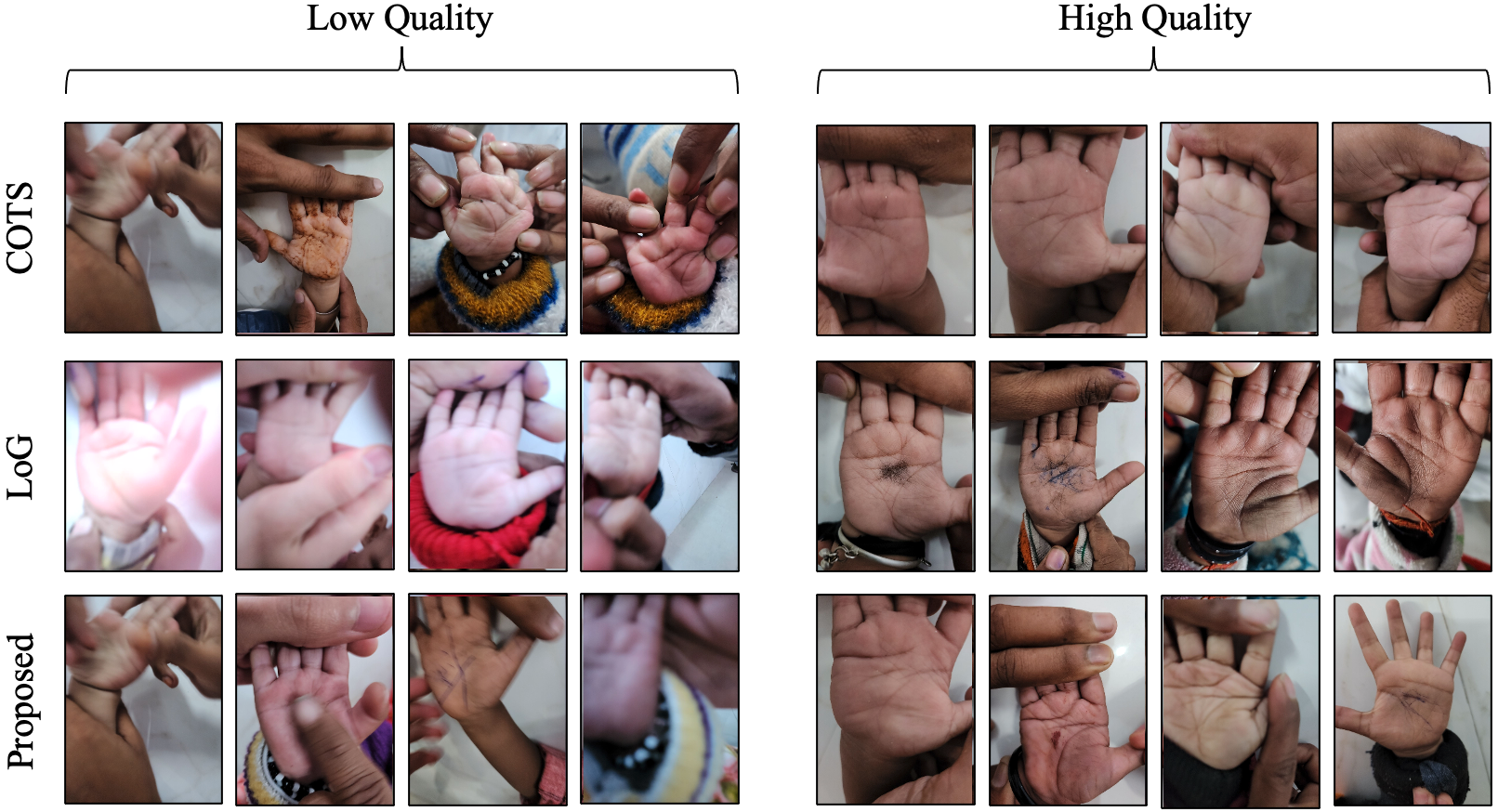}
\caption{Low and high quality images predicted by Armatura SDK (top row), variance of LoG (middle row) and the proposed (bottom row) quality metrics. Both the Armatura SDK and the proposed quality metric incorporate factors such as occlusion, blurriness, and perspective degradation; whereas the variance of LoG metric is only a proxy for image sharpness, as it only measures the prevalence of edges in the image.}
\label{fig:quality_vis}
\end{center}
\vspace{-2em}
\end{figure*}

Lastly, we introduced a palmprint quality estimation metric to identify low quality images. In particular, the L2 norm of the Palm-ID embedding is used as the quality value, providing a more accurate estimate of the objective quality of the image compared to the quality metric implemented by Armatura's SDK and the variance of the Laplacian of Gaussian (a common method for estimating blur in an image). The error-reject trade-off curve in Figure~\ref{fig:ert} demonstrates the quantitative advantages of the L2 norm over the other methods. A visual comparison of low and high-quality predictions by Armatura, the variance of the Laplacian of Gaussian, and our method is presented in Figure~\ref{fig:quality_vis}. 

The variance of the Laplacian of Gaussian (LoG) metric was computed on the Region of Interests (ROIs) rather than the full palm images, as this was found to be less susceptible to background noise. Furthermore, we crop 32 pixels from all four sides of the ROI to ensure that the LoG is not influenced by minor occlusions, such as the edges of shirt sleeves, etc. For Armatura's metric, we directly use the quality scores output by the system.


In most datasets, our quality metric performs the best, except for CASIA and IITD. In the case of IITD, there are only a few mis-classifications so the difference in performance among the three metrics seems to be negligible. However, in the case of CASIA, the poor quality samples seem to primarily stem from a few poor contrast, blurry samples, which the LoG metric is primarily designed to detect.

\section{Conclusion}
In conclusion, this research significantly advances contactless palmprint recognition by addressing inherent challenges through a multifaceted approach. The integration of global and local features, ViT-CNN fusion, and a novel palmprint enhancement module collectively contribute to improved matching accuracy on a variety of public and private databases collected by different sensors. The efficiency of the system is enhanced through non-linear dimensionality reduction, enabling deployment on mobile devices without compromising accuracy. Cross-database and time-separated evaluations demonstrate the robustness and practical applicability of the proposed method. The research contributes to the field by presenting a comprehensive end-to-end pipeline, showcasing the feasibility of on-device recognition processes, and providing a valuable resource for future advancements in contactless palmprint recognition technology on commodity smartphones.

\ifCLASSOPTIONcaptionsoff
  \newpagegod
\fi

\bibliography{cite}

\begin{thebibliography}{10}

\bibitem{jainbiometrics2011}
A.~K. Jain, A.~A. Ross, and K.~Nandakumar, {\em Introduction to Biometrics}.
\newblock Springer Publishing Company, 2011.

\bibitem{grosz2021c2cl}
S.~A. Grosz, J.~J. Engelsma, E.~Liu, and A.~K. Jain, ``{C2CL}: Contact to {C}ontactless {F}ingerprint {M}atching,'' {\em IEEE Trans. IFS}, vol.~17, 2021.

\bibitem{dabouei2019deep}
A.~Dabouei, S.~Soleymani, J.~Dawson, and N.~M. Nasrabadi, ``Deep contactless fingerprint unwarping,'' in {\em 2019 International Conference on Biometrics (ICB)}, pp.~1--8, IEEE, 2019.

\bibitem{dong2023synthesis}
C.~Dong and A.~Kumar, ``Synthesis of multi-view 3d fingerprints to advance contactless fingerprint identification,'' {\em IEEE Transactions on Pattern Analysis and Machine Intelligence}, 2023.

\bibitem{herschel1916origin}
W.~J. Herschel, {\em {The Origin of Finger-Printing}}.
\newblock H. Milford, Oxford University Press, 1916.

\bibitem{armatura}
Armatura, ``{PalmMobileSDK}.''
\newblock https://armatura.us/PalmMobileSDK/.

\bibitem{Yang_2022}
Z.~Yang, ``{Tencent wants you to pay with your palm. What could go wrong?},'' Nov 2022.
\newblock https://www.technologyreview.com/2022/11/15/1063218/whats-next-biometrics-palm-print-recognition-tencent-we-chat-pay/.

\bibitem{amazon}
Amazon, ``{Amazon One},'' 2021.
\newblock https://one.amazon.com/.

\bibitem{redrock}
``{PalmID - Palmprint Biometrics for Devices with Camera},'' May 2023.
\newblock https://www.redrockbiometrics.com/.

\bibitem{fujitsu}
``{Palm Vein Pattern Authentication Technology},'' 2020.
\newblock https://www.fujitsu.com/downloads/COMP/ffna/palm-vein/palmsecure\_wp.pdf.

\bibitem{7754902}
M.~M.~H. Ali, P.~Yannawar, and A.~T. Gaikwad, ``{Study of Edge Detection Methods Based on Palmprint Lines},'' in {\em 2016 International Conference on Electrical, Electronics, and Optimization Techniques (ICEEOT)}, pp.~1344--1350, 2016.

\bibitem{lu2003palmprint}
G.~Lu, D.~Zhang, and K.~Wang, ``{Palmprint Recognition Using Eigenpalms Features},'' {\em Pattern Recognition Letters}, vol.~24, no.~9-10, pp.~1463--1467, 2003.

\bibitem{connie2005automated}
T.~Connie, A.~T.~B. Jin, M.~G.~K. Ong, and D.~N.~C. Ling, ``{An automated Palmprint Recognition System},'' {\em Image and Vision computing}, vol.~23, no.~5, pp.~501--515, 2005.

\bibitem{zhang2017towards}
L.~Zhang, L.~Li, A.~Yang, Y.~Shen, and M.~Yang, ``Towards {C}ontactless {P}almprint {R}ecognition: {A} {N}ovel {D}evice, a {N}ew {B}enchmark, and a {C}ollaborative {R}epresentation {B}ased {I}dentification {A}pproach,'' {\em Pattern Recognition}, vol.~69, pp.~199--212, 2017.

\bibitem{sunordinal}
Z.~Sun, T.~Tan, Y.~Wang, and S.~Li, ``Ordinal {P}almprint {R}epresentation for {P}ersonal {I}dentification,'' in {\em Proceedings of the IEEE CVPR}, 2005.

\bibitem{trabelsi2022efficient}
S.~Trabelsi, D.~Samai, F.~Dornaika, A.~Benlamoudi, K.~Bensid, and A.~Taleb-Ahmed, ``{Efficient Palmprint Biometric Identification Systems Using Deep Learning and Feature Selection Methods},'' {\em Neural Computing and Applications}, vol.~34, no.~14, pp.~12119--12141, 2022.

\bibitem{godbole2023child}
A.~Godbole, S.~A. Grosz, and A.~K. Jain, ``{Child Palm-ID: Contactless Palmprint Recognition for Children},'' {\em arXiv preprint arXiv:2305.05161}, 2023.

\bibitem{yulin2023best}
F.~Yulin and A.~Kumar, ``{BEST: Building Evidences From Scattered Templates for Accurate Contactless Palmprint Recognition},'' {\em Pattern Recognition}, vol.~138, p.~109422, 2023.

\bibitem{liang2022innovative}
X.~Liang, Z.~Li, D.~Fan, B.~Zhang, G.~Lu, and D.~Zhang, ``{Innovative Contactless Palmprint Recognition System Based on Dual-Camera Alignment},'' {\em IEEE Transactions on Systems, Man, and Cybernetics: Systems}, vol.~52, no.~10, pp.~6464--6476, 2022.

\bibitem{zhao2020deep}
S.~Zhao and B.~Zhang, ``{Deep Discriminative Representation for Generic Palmprint Recognition},'' {\em Pattern Recognition}, vol.~98, p.~107071, 2020.

\bibitem{matkowski2019palmprint}
W.~M. Matkowski {\em et~al.}, ``Palmprint {R}ecognition in {U}ncontrolled and {U}ncooperative {E}nvironment,'' {\em IEEE Trans. IFS}, 2019.

\bibitem{vaswani2017attention}
A.~Vaswani, N.~Shazeer, N.~Parmar, J.~Uszkoreit, L.~Jones, A.~N. Gomez, {\L}.~Kaiser, and I.~Polosukhin, ``Attention is all you need,'' {\em Advances in Neural Information Processing Systems}, vol.~30, 2017.

\bibitem{tandon2022transformer}
S.~Tandon and A.~Namboodiri, ``Transformer based fingerprint feature extraction,'' in {\em 2022 26th International Conference on Pattern Recognition (ICPR)}, pp.~870--876, IEEE, 2022.

\bibitem{zhong2021face}
Y.~Zhong and W.~Deng, ``Face transformer for recognition,'' {\em arXiv preprint arXiv:2103.14803}, 2021.

\bibitem{delgado2023exploring}
P.~Delgado-Santos, R.~Tolosana, R.~Guest, F.~Deravi, and R.~Vera-Rodriguez, ``Exploring transformers for behavioural biometrics: A case study in gait recognition,'' {\em Pattern Recognition}, vol.~143, p.~109798, 2023.

\bibitem{raghu2021vision}
M.~Raghu, T.~Unterthiner, S.~Kornblith, C.~Zhang, and A.~Dosovitskiy, ``Do vision transformers see like convolutional neural networks?,'' {\em Advances in Neural Information Processing Systems}, vol.~34, pp.~12116--12128, 2021.

\bibitem{grosz2022minutiae}
S.~A. Grosz, J.~J. Engelsma, R.~Ranjan, N.~Ramakrishnan, M.~Aggarwal, G.~G. Medioni, and A.~K. Jain, ``Minutiae-guided fingerprint embeddings via vision transformers,'' {\em arXiv preprint arXiv:2210.13994}, 2022.

\bibitem{grosz2023afr}
S.~A. Grosz and A.~K. Jain, ``Afr-net: Attention-driven fingerprint recognition network,'' {\em IEEE Transactions on Biometrics, Behavior, and Identity Science}, 2023.

\bibitem{kingma2014adam}
D.~P. Kingma and J.~Ba, ``Adam: A method for stochastic optimization,'' {\em arXiv preprint arXiv:1412.6980}, 2014.

\bibitem{kumar2008incorporating}
A.~Kumar, ``{Incorporating Cohort Information for Reliable Palmprint Authentication},'' in {\em Sixth ICVGIP}, IEEE, 2008.

\bibitem{hao2008multispectral}
Y.~Hao, Z.~Sun, T.~Tan, and C.~Ren, ``Multispectral {P}alm {I}mage {F}usion for {A}ccurate {C}ontact-{F}ree {P}almprint {R}ecognition,'' in {\em 15th IEEE ICIP}, 2008.

\bibitem{jain2008latent}
A.~K. Jain and J.~Feng, ``{Latent Palmprint Matching},'' {\em IEEE Transactions on Pattern Analysis and Machine Intelligence}, vol.~31, no.~6, pp.~1032--1047, 2008.

\bibitem{cappelli2012fast}
R.~Cappelli, M.~Ferrara, and D.~Maio, ``A fast and accurate palmprint recognition system based on minutiae,'' {\em IEEE Transactions on Systems, Man, and Cybernetics, Part B (Cybernetics)}, vol.~42, no.~3, pp.~956--962, 2012.

\bibitem{grosz2023latent}
S.~A. Grosz and A.~K. Jain, ``Latent fingerprint recognition: Fusion of local and global embeddings,'' {\em IEEE Transactions on Information Forensics and Security}, vol.~18, pp.~5691--5705, 2023.

\bibitem{rombach2022high}
R.~Rombach, A.~Blattmann, D.~Lorenz, P.~Esser, and B.~Ommer, ``High-resolution image synthesis with latent diffusion models,'' in {\em Proceedings of the IEEE/CVF conference on computer vision and pattern recognition}, pp.~10684--10695, 2022.

\bibitem{von-platen-etal-2022-diffusers}
P.~von Platen, S.~Patil, A.~Lozhkov, P.~Cuenca, N.~Lambert, K.~Rasul, M.~Davaadorj, and T.~Wolf, ``Diffusers: State-of-the-art diffusion models.'' \url{https://github.com/huggingface/diffusers}, 2022.

\bibitem{loshchilov2017decoupled}
I.~Loshchilov and F.~Hutter, ``Decoupled weight decay regularization,'' {\em arXiv preprint arXiv:1711.05101}, 2017.

\bibitem{deng2019arcface}
J.~Deng, J.~Guo, N.~Xue, and S.~Zafeiriou, ``Arcface: Additive angular margin loss for deep face recognition,'' in {\em Proceedings of the IEEE/CVF conference on computer vision and pattern recognition}, pp.~4690--4699, 2019.

\bibitem{engelsma2022hers}
J.~J. Engelsma, A.~K. Jain, and V.~N. Boddeti, ``{HERS: Homomorphically Encrypted Representation Search},'' {\em IEEE Transactions on Biometrics, Behavior, and Identity Science}, vol.~4, no.~3, pp.~349--360, 2022.

\bibitem{gong2019intrinsic}
S.~Gong, V.~N. Boddeti, and A.~K. Jain, ``{On the Intrinsic Dimensionality of Image Representations},'' in {\em Proceedings of the IEEE/CVF Conference on Computer Vision and Pattern Recognition}, pp.~3987--3996, 2019.

\bibitem{kim2022adaface}
M.~Kim, A.~K. Jain, and X.~Liu, ``Adaface: Quality adaptive margin for face recognition,'' in {\em Proceedings of the IEEE/CVF conference on computer vision and pattern recognition}, pp.~18750--18759, 2022.

\bibitem{afifi201911kHands}
M.~Afifi, ``{11K Hands: Gender Recognition and Biometric Identification Using a Large Dataset of Hand Images},'' {\em Multimedia Tools and Applications}, 2019.

\bibitem{aykut2015developing}
M.~Aykut and M.~Ekinci, ``{Developing a Contactless Palmprint Authentication System by Introducing a Novel ROI Extraction Method},'' {\em Image and Vision Computing}, vol.~40, pp.~65--74, 2015.

\bibitem{izadpanahkakhk2019novel}
M.~Izadpanahkakhk {\em et~al.}, ``{Novel Mobile Palmprint Databases for Biometric Authentication},'' {\em International Journal of Grid and Utility Computing}, vol.~10, no.~5, pp.~465--474, 2019.

\bibitem{coep}
COEP, ``{COEP} {P}alm {P}rint {D}atabase.''
\newblock https://www.coep.org.in/resources/coeppalmprintdatabase.

\bibitem{hao2007comparative}
Y.~Hao, Z.~Sun, and T.~Tan, ``{Comparative Studies on Multispectral Palm Image Fusion for Biometrics},'' in {\em Asian conference on computer vision}, pp.~12--21, Springer, 2007.

\bibitem{kumarpersonal}
A.~Kumar and S.~Shekhar, ``{Personal Identification Using Rank-level Fusion},'' {\em IEEE Trans. Systems, Man, and Cybernetics: Part C}, pp.~743--752, 2011.

\end{thebibliography}
\bibliographystyle{ieeetr}

\vspace{-3em}
\begin{IEEEbiography}[{\includegraphics[width=1in,height=1.25in,clip,keepaspectratio]{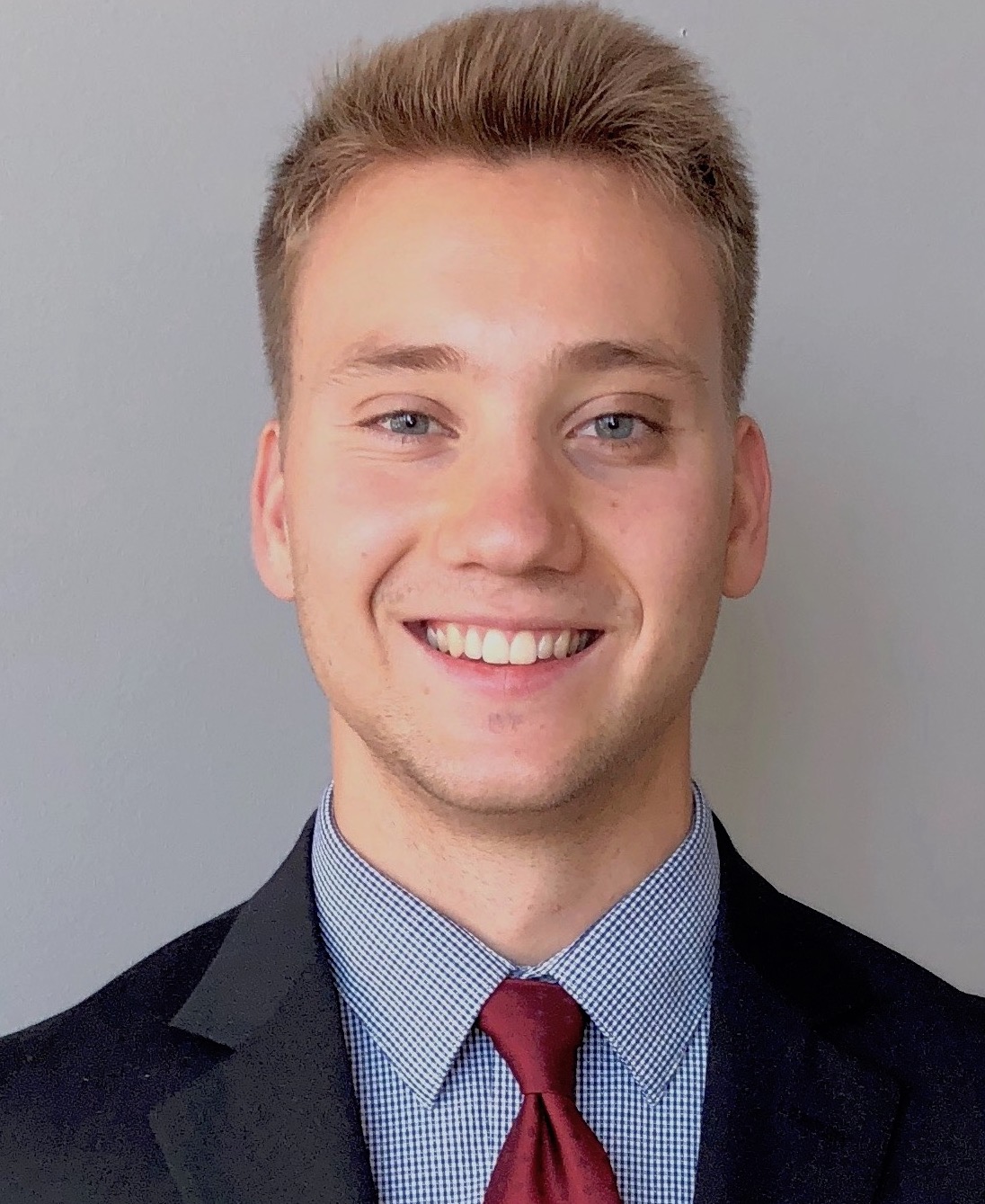}}]{Steven A. Grosz}
received his B.S. degree with highest honors in Electrical Engineering from Michigan State University, East Lansing, Michigan, in 2019. He is currently a doctoral student in the Department of Computer Science and Engineering at Michigan State University. His primary research interests are in the areas of machine learning and computer vision with applications in biometrics.
\end{IEEEbiography}
\vspace{-3em}

\begin{IEEEbiography}[{\includegraphics[width=1in,height=1.25in,clip,keepaspectratio]{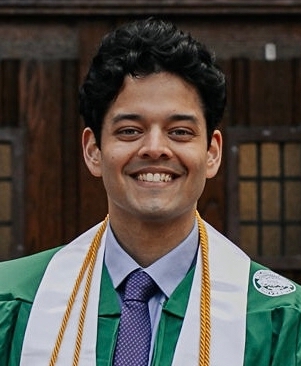}}]{Akash Godbole}
received his B.S. degree with honors in Computer Science from Michigan State University, East Lansing, Michigan, in 2021. He is currently a graduate student in the Department of Computer Science and Engineering at Michigan State University. His research interests include biometrics, machine learning, and computer vision.
\end{IEEEbiography}
\vspace{-3em}

\begin{IEEEbiography}[{\includegraphics[width=1in,height=1.25in,clip,keepaspectratio]{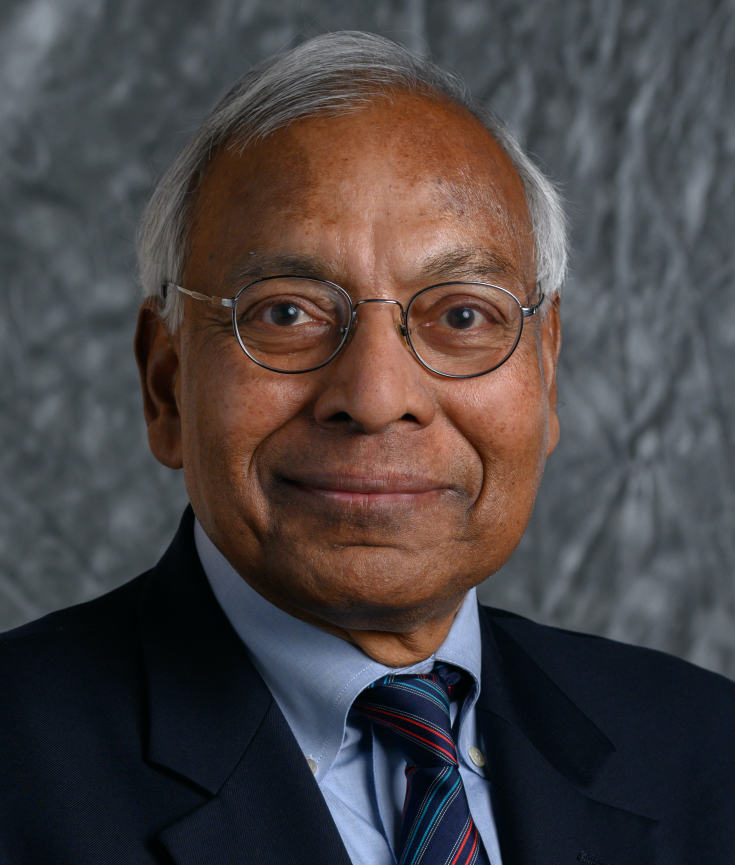}}]{Anil K. Jain}
Anil K. Jain is a University distinguished professor in the Department of Computer Science and Engineering at Michigan State University. His research interests include pattern recognition, computer vision, and biometric authentication. He served as the editor-in-chief of the IEEE Transactions on Pattern Analysis and Machine Intelligence and was a member of the United States Defense Science Board. He has received Fulbright, Guggenheim, Alexander von Humboldt, and IAPR King Sun Fu awards. He is an elected member of the National Academy of Engineering, the Indian National Academy of Engineering, the World Academy of Sciences, and the Chinese Academy of Sciences.
\end{IEEEbiography}

\end{document}